\documentclass[11pt]{article}

\usepackage[final]{acl}

\usepackage{times}
\usepackage{latexsym}
\usepackage{booktabs}
\usepackage{amssymb}
\usepackage{amsmath}
\usepackage{adjustbox}

\usepackage[T1]{fontenc}

\usepackage[utf8]{inputenc}
\usepackage{multirow}

\usepackage{microtype}

\usepackage{inconsolata}
\usepackage[table]{xcolor}
\usepackage{graphicx}
\usepackage{tabularx}
\usepackage{colortbl}
\usepackage{xcolor}
\usepackage{float}

\usepackage{longtable}
\usepackage{array}
\usepackage{enumitem}

\title{LENS: LLM-Enabled Narrative Synthesis for Mental Health by Aligning Multimodal Sensing with Language Models}

\author{
  \textbf{Wenxuan Xu\textsuperscript{*1}},
  \textbf{Arvind Pillai\textsuperscript{*1}},
  \textbf{Subigya Nepal\textsuperscript{2}},
  \textbf{Amanda C Collins\textsuperscript{3, 4}},
  \\
  \textbf{Daniel M Mackin\textsuperscript{1}},
  \textbf{Michael V Heinz\textsuperscript{1}},
  \textbf{Tess Z Griffin\textsuperscript{1}},
  \textbf{Nicholas C Jacobson\textsuperscript{1}},
  \\
  \textbf{Andrew Campbell \textsuperscript{1}},
  \\
  \textsuperscript{1}Dartmouth College,
  \textsuperscript{2}University of Virginia,
  \\
  \textsuperscript{3}Massachusetts General Hospital,
  \textsuperscript{4}Harvard Medical School,
  \\
  \textsuperscript{*}Equal contribution,
  \\
  \small{
    \textbf{Correspondence:} \texttt{\{wenxuan.xu.gr, arvind.pillai.gr\}@dartmouth.edu}
  }
}

\begin{document}
\maketitle
\begin{abstract}

  Multimodal health sensing offers rich behavioral signals for assessing mental health, yet translating these numerical time-series measurements into natural language remains challenging. Current LLMs cannot natively ingest long-duration sensor streams, and paired sensor–text datasets are scarce. To address these challenges, we introduce LENS, a framework that aligns multimodal sensing data with language models to generate clinically grounded mental-health narratives. LENS first constructs a large-scale dataset by transforming Ecological Momentary Assessment (EMA) responses related to depression and anxiety symptoms into natural-language descriptions, yielding over 100,000 sensor–text QA pairs from 258 participants. To enable native time-series integration, we train a patch-level encoder that projects raw sensor signals directly into an LLM’s representation space. Our results show that LENS outperforms strong baselines on standard NLP metrics and task-specific measures of symptom-severity accuracy. A user study with 13 mental-health professionals further indicates that LENS-produced narratives are comprehensive and clinically meaningful. Ultimately, our approach advances LLMs as interfaces for health sensing, providing a scalable path toward models that can reason over raw behavioral signals and support downstream clinical decision-making~\footnote{Code available at \url{https://github.com/Wen-xuan-Xu/LENS}}.

\end{abstract}

\section{Introduction}

Mental health conditions on the spectrum of anxiety and depression affect an estimated 18\% and 9.5\% of adults in the United States each year~\citep{hopkins_mental_health}. Traditional screening methods typically rely on structured clinical interviews and validated self-report instruments such as the Patient Health Questionnaire (PHQ-9)~\citep{kroenke2001phq} and the Generalized Anxiety Disorder scale (GAD-2)~\citep{spitzer2006brief}. However, these assessments are limited by their high burden on clinicians, dependence on retrospective self-reports, and reduced ecological validity because they are administered in controlled settings that do not capture an individual’s real-world context~\citep{abd2023systematic}.

\begin{figure}
  \centering
  \includegraphics[width=1\linewidth]{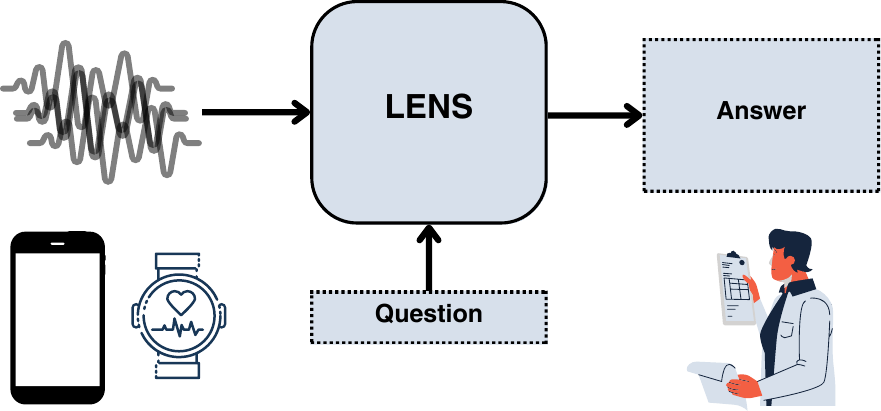}
  \caption{\textbf{Illustration of the LENS idea.} Given sensor signals from mobile and wearable devices, LENS takes a question as input and produces a natural-language description. Clinicians can then view an interpretable snapshot of the user's mental state instead of raw sensor streams.}
  \label{fig:teaser}
  \vspace{-0.5cm}
\end{figure}

To address these challenges, recent work has used mobile and wearable technologies to collect passive sensing data (for example, phone usage, speech features, and heart rate) and to administer ecological momentary assessments (EMA)~\citep{GLOBEM, gomes2023survey, nepal2024capturing, nepal2024social}. A growing body of evidence shows that behavioral and physiological signals such as activity levels, sleep patterns, mobility, voice characteristics, and smartphone interactions can indicate the severity of depression and anxiety symptoms~\citep{wang2014studentlife, sheikh2021wearable, jacobson2021deep, pillai2023rare}. Together, these findings highlight passive sensing as a promising complementary approach for mental health monitoring.

In parallel, recent work has demonstrated the potential of large language models (LLMs) for mental health assessment. Studies show that prompt engineering and fine-tuning can enable depression detection, symptom severity inference, physiological indicator prediction, and even generation of psychological rationales~\cite{YangMentaLLaMA2024, moon2025depressllm, kim2024healthllm}. Despite these promising capabilities, LLMs struggle with long-duration time series due to limitations in context length and tokenizer design, which prevent them from directly ingesting raw numerical sequences~\cite{spathis2024first}. In contrast, multimodal vision and language models benefit from mature frameworks that support native visual understanding~\cite{li2022blip, dosovitskiy2021ViT, radford2021Clip}. However, comparable methods for native time-series integration remain scarce~\cite{tan2024language}. As a result, the few studies that combine passive health sensing with LLMs rarely operate directly on raw sensor streams~\cite{kim2024healthllm, justin2024towards, englhardt_classification_2024}. Overall, progress in integrating behavioral sensing data with language is limited by the lack of large datasets that pair raw sensor streams with text and by the limited capabilities of existing methods that align time-series signals with language models.

We envision LENS as a \textit{pre-consultation monitoring tool} for individuals already receiving clinical care for depression or anxiety, rather than a population-level screening or diagnostic instrument. In this intended workflow, clinicians receive a naturalistic behavioral summary of a patient's recent state derived from passive sensing, enabling more informed and time-efficient consultations~\ref{fig:teaser}. Toward this goal, our contributions are threefold. \textbf{(1) LENS data synthesis pipeline:} We introduce a pipeline that generates semantic mental health descriptions of multimodal sensing data from EMA responses (Section \ref{sec:lens_dataset}), producing a dataset of more than 100,000 sensor–text pairs and directly addressing the shortage of such resources.
\textbf{(2) LENS training:} We propose a training strategy based on a patch-level time-series encoder that projects sensor signals into the language model’s representation space (Section \ref{sec:lens_training}). By interleaving time-series and text embeddings and using a two-stage curriculum, we show that LENS can generate clinically grounded narratives.
\textbf{(3) Comprehensive evaluation:} We evaluate our approach on a clinical dataset of 258 participants comprising 50,957 unique samples (Section \ref{sec:results}). Beyond custom LLM metrics, we conduct a user study with 13 mental health experts who manually assessed 117 narratives.

\section{Related Work} \label{sec:related_work}

Recent work explores mobile and wearable data for mental health prediction, primarily focusing on symptom classification \citep{kim2024healthllm, englhardt_classification_2024} or improving reasoning via multimodal encoders \citep{justin2024towards}. These systems prioritize prediction over natural language generation, often producing text as secondary reasoning traces. Furthermore, methods that serialize numerical data into tokens face significant scalability and tokenization constraints~\cite{pillai_beyond_2025, spathis2024first, yoon2024my}, and deploying models on heterogeneous edge devices introduces additional challenges around data distribution and communication efficiency~\citep{liu2025feddhad}. In contrast, LENS anchors narrative generation in raw sensor measurements and clinically validated PHQ/GAD items to produce meaningful symptom descriptions.

LLMs are increasingly used to synthesize paired datasets for time-series tasks by generating artificial sequences or surrogate signals for QA pairs and explanations \citep{xie_chatts_2025, li_sensorllm_2025, yan_voila-_nodate, imran_llasa_2024}. While scalable, these pipelines rely on synthetic inputs that fail to capture the complexity of real physiological and behavioral signals; long-tail distributional skew in generated outputs further complicates quality assurance~\citep{zhou2025april}. LENS instead pairs real-world sensor streams with clinical assessments, using LLMs only to refine linguistic fluency via rigorous quality checks.

While multimodal alignment has progressed rapidly across vision--language understanding \citep{liu2023visualinstructiontuning}, embodied view synthesis \citep{wang2025holigs}, and video-based physical reasoning \citep{liu2026mosiv}, comparable integration for time-series data remains limited. Existing text-based serialization is constrained by numerical encoding and context length \citep{xue2023promptcast, gruver2023large}, while vision-based methods convert series into images, introducing plot-engineering biases and indirect representations \citep{yoon2024my, zhang2023insight}. Alignment-based approaches project encoder representations into LLM hidden states \citep{ming2023timellm, xie_chatts_2025}. However, by optimizing for synthetic QA, they rarely generate natural language aligned with real-world clinical constructs, and the reliability of standard benchmarks in specialized domains remains an open concern~\citep{yao2025measure}. Our work differs from these approaches by producing symptom-oriented narratives tied to psychometric instruments and grounded directly in raw signals

\section{Methods}
LENS consists of two components: a scalable dataset-construction pipeline that transforms EMA responses into high-quality sensor–text pairs (Section \ref{sec:lens_dataset}), and a sensor–text alignment method that enables native integration of raw time-series signals into an LLM through a patch-based encoder (Section \ref{sec:lens_training}).

\subsection{LENS: Dataset Construction} \label{sec:lens_dataset}

\subsubsection{Study} Our longitudinal study investigates intra-day fluctuations in mental health symptoms among individuals diagnosed with major depressive disorder. In this 90-day study, we recruited participants aged 18 years and older, residing in the United States. Each participant wore a Garmin vivoactive 3 device and installed our Android application. This setup enabled the collection of passive sensing data, which are behavioral and physiological signals automatically recorded from smartphones and wearables, and ecological momentary assessments (EMAs), in which participants actively reported depression and anxiety symptoms experienced over the past four hours. Non-identifiable study data will be available at the NIMH Data Archive (NDA)\footnote{\url{https://nda.nih.gov}}. Researchers requesting access must submit a Data Use Certification (DUC) through the NDA platform, which requires institutional sponsorship and the signature of an Authorized Institutional Business Official at an institution with an active Federal Wide Assurance.

The EMA consists of 14 items adapted from the PHQ-9 and GAD-4 (Table \ref{tab:ema_question_categories}; Appendix \S\ref{apd:ema_questions}). It is administered three times per day in the morning, afternoon, and evening, customized to each participant’s waking time. Each item is rated on a continuous 0 (“Not at all”) to 100 (“Constantly”) scale. In addition to the EMAs, we collect mobile and wearable time-series signals, including GPS traces, step counts, accelerometer-derived zero-crossing rate (ZCR) and energy, conversation time, phone lock and unlock events, heart rate, sleep estimates, and stress levels. These signals have been shown to correlate with depressive and anxiety symptoms~\cite{choudhary2022machineML}. To align self-reported EMAs with corresponding sensing data, we use each EMA’s completion time to retrieve the preceding four hours of sensor data. This procedure results in a temporally aligned, multi-modal dataset suitable for narrative synthesis and modeling. Ultimately, we utilize 50,957 EMAs from 258 participants. Although all participants carry an MDD diagnosis, the ordinal EMA distributions span the full 0--3 severity range across all nine PHQ-9 items, with over 42\% of samples falling into the ``No/Minimal'' or ``Mild'' categories, ensuring sufficient within-cohort variability (Table~\ref{tab:phq9_distribution}).

\subsubsection{Sensor Data Processing} The sensing data includes two types of signals: (1) continuous streams, which contain time-series data such as steps, heart rate, accelerometer-derived zero-crossing rate (ZCR) and energy, phone lock and unlock state, stress level, and GPS traces, and (2) aggregated streams, which contain daily or window-level values such as sleep duration from the previous night and total conversation time. After preprocessing, all signals are standardized to fixed sampling rates to ensure consistent temporal alignment across modalities (see Appendix~\S\ref{apd:sensor_preprocessing} for specific details).

\begin{figure}[t]
  \centering
  \includegraphics[width=1\linewidth]{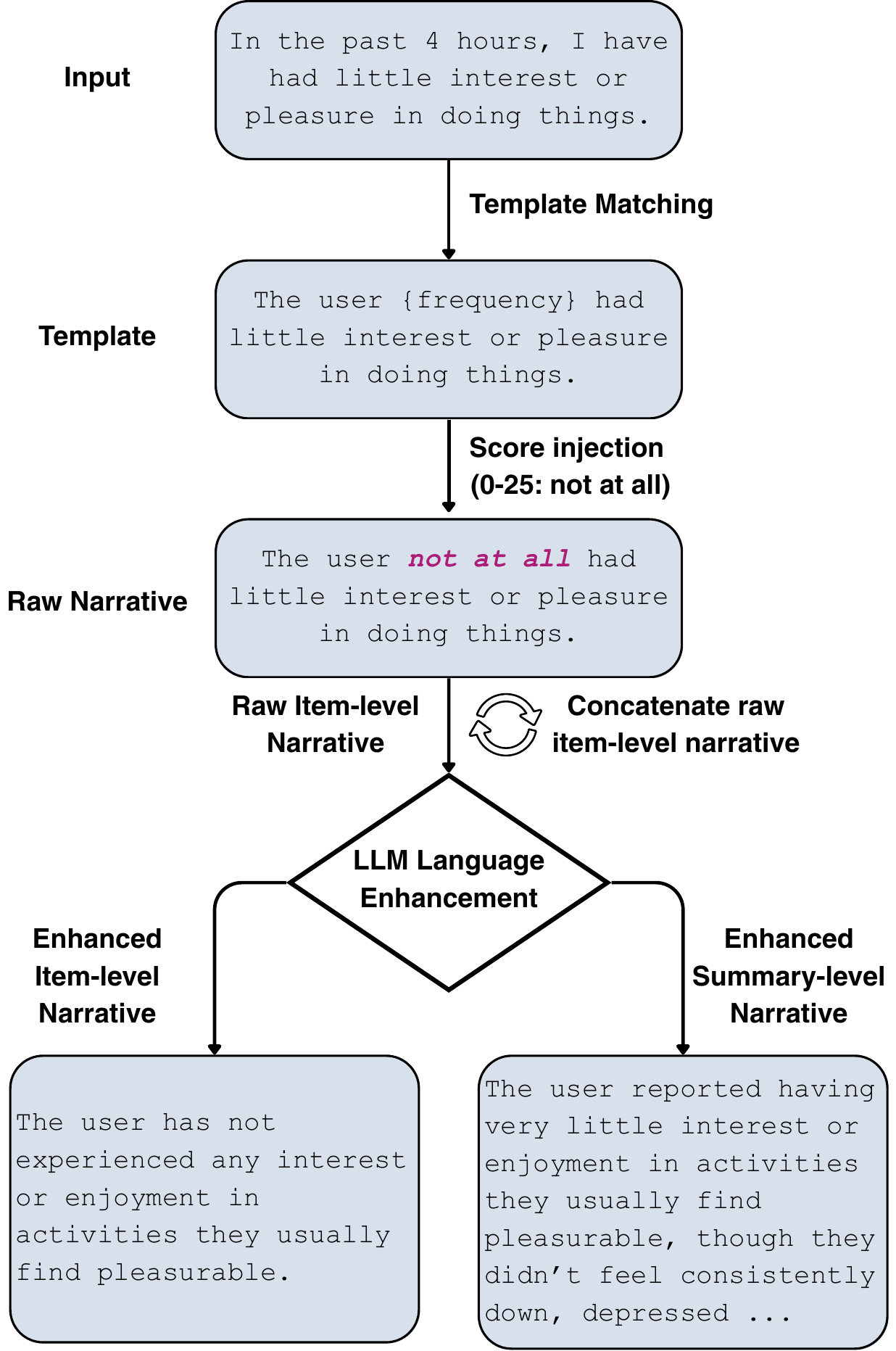}
  \caption{\textbf{Overview of the narrative synthesis pipeline.} EMA questions and responses are mapped to templates populated with corresponding frequency phrases. Subsequently, an LLM refines the text at both the item and summary levels (concatenated narratives) to enhance fluency and lexical diversity.}
  \label{fig:data_synthesis_example}
  \vspace{-0.25cm}
\end{figure}

\begin{figure*}
  \centering
  \includegraphics[width=1\linewidth]{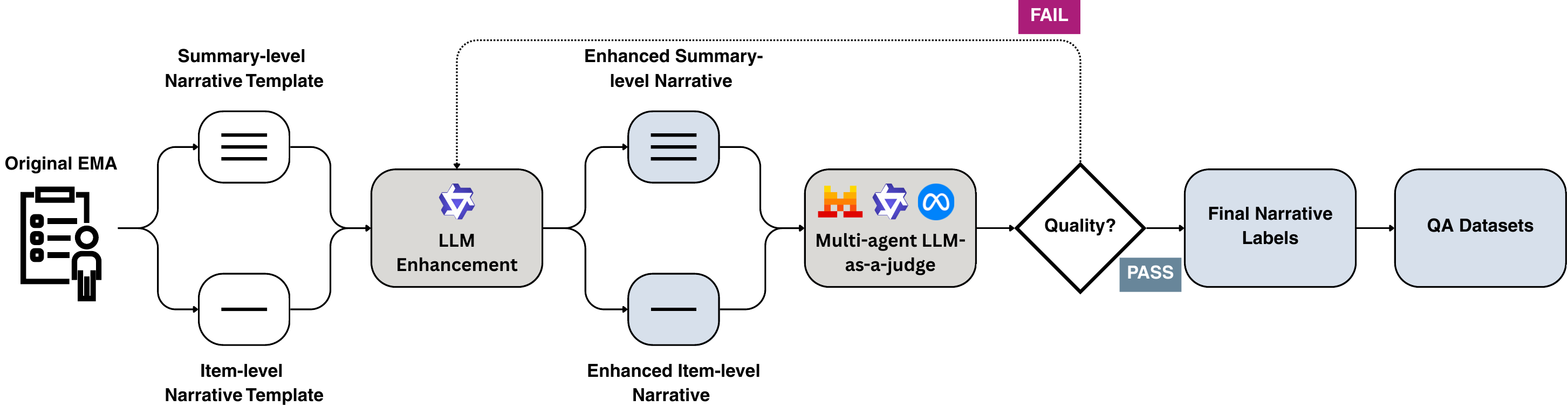}
  \caption{\textbf{LENS dataset construction pipeline.} EMA responses are first converted into item-level and summary template narratives, which are then rewritten into fluent, enhanced narratives using Qwen2.5-14B. A multi-agent LLM-as-a-judge system conducts automatic quality control, routing failed cases back for regeneration until they satisfy all criteria. The final accepted narratives are then paired with paraphrased question variants to construct the QA datasets.}
  \label{fig:lens_dataset_construction}
\end{figure*}

\subsubsection{LENS Narrative Synthesis} \label{sec:lens_narrative_synthesis}To address the lack of sensor-text datasets describing mental health symptoms, LENS constructs high-quality QA datasets from self-reported EMA responses to serve as ground truth for model training and evaluation. Because each numeric EMA response corresponds to a symptom intensity category aligned with DSM-5 semantics, these values can be converted into natural language descriptions of participants’ experiences. We produce two QA datasets: (1) \textbf{Item-level QA dataset}: Each sample represents a single EMA item and supports question-specific supervision, and (2) \textbf{Summary-level QA dataset:} Each sample reflects all EMA questions and targets full-summary generation.

\textbf{Responses $\rightarrow$ Answers (A)}. EMA responses are converted into narrative labels as shown in Figure \ref{fig:data_synthesis_example}. For each question, we first transform the EMA item into a symptom-focused template sentence. We then insert the frequency phrase associated with its numeric range (0–25: not at all, 26–50: sometimes, 51–75: often, 76–100: constantly) to obtain the raw item-level narrative. Question 14 is treated as binary, and overall severity is categorized as mild, moderate, or severe. Summary-level narratives are formed by concatenating item-level narratives before applying LLM refinement. To reduce the mechanical tone of the templates, we use prompt-based rewriting with the locally deployed Qwen2.5-14B model. The system prompt enforces factual accuracy, consistent terminology, and stigma-free language, while the user prompt supplies the rule-based text and requests a fluent rewrite (See prompt template in Appendix~\S\ref{apd:prompt_templates}). This process produces 101,914 item-level narratives and 50,957 summary-level narratives that serve as ground truth for training LENS.

\textbf{Questions (Q).} The ground-truth questions consist of item-level prompts for each EMA question and a summary-level prompt that covers all items along with the overall severity statement. To increase linguistic diversity, we use GPT-4o to generate paraphrased variants of every question. For each original prompt, we created ten semantically equivalent but lexically distinct phrasings. During QA dataset construction, one paraphrased variant is randomly sampled for each instance.

\textbf{Quality Control.} To ensure narrative reliability, we implement an automatic quality-control pipeline using a multi-model LLM-judge system (Figure \ref{fig:lens_dataset_construction}). Each judge model follows an LLM-as-a-Judge prompting template~\cite{li_generation_2025} with rule-augmented instructions that embed evaluation principles, baseline references, and dimension-specific rubrics directly in the prompt. Judges compare the template-based and enhanced narratives in a pairwise format and output five dimension scores (1–5), confidence values, and a short rationale, providing structured and transparent assessments. Because individual LLM judges can be biased, we use three independent models (Mistral-7B~\cite{mistral7b}, Llama-3.1-8B~\cite{llama3}, and Qwen2.5-7B~\cite{qwen2025qwen25technicalreport}). Dimension scores are averaged and rounded, and their sum forms a total quality score; confidence values are averaged similarly. A narrative is accepted only if its average total score exceeds 20 (out of 25) and its mean confidence exceeds 0.8. Otherwise, it is returned for regeneration and re-evaluation (FAIL in Figure \ref{fig:lens_dataset_construction}). This iterative refine-and-judge loop, akin to Refine-n-Judge~\cite{cayir2025refinenjudge}, filters out low-quality outputs and ensures that finalized narratives faithfully reflect EMA responses while improving fluency and diversity without introducing distortions. Importantly, the multi-stage LLM pipeline is used strictly for offline dataset construction and is never invoked at inference time; the deployed model consists solely of the patch-based encoder and the LLM backbone. Using commercial API pricing as a conservative upper bound, the total equivalent cost of constructing the full dataset is approximately \$7.75 (\$0.00015 per sample), ensuring high scalability (Table~\ref{tab:cost_analysis}).

\subsection{LENS: Training \& Architecture} \label{sec:lens_training}

\subsubsection{Time-series Encoder}
We encode each time–series stream independently using a lightweight patch-based module that converts raw scalar values into language-model-compatible embeddings. Given a univariate sequence $S=\{s_t\}_{t=1}^{T}$, we first apply a reversible value normalization to stabilize scale while keeping absolute magnitudes recoverable. Let $\mu$ and $\sigma$ denote the per-stream mean and standard deviation, then
$\tilde{s}_t=\frac{s_t-\mu}{\sigma}$, and auxiliary statistics such as $(\mu,\sigma,m_{\min},m_{\max})$ are inserted into the textual prompt as metadata, allowing the model to reason about original numerical ranges inside the language space~\citep{opentslm,xie_chatts_2025}. The normalized sequence is divided into $N=\lceil T/w \rceil$ non-overlapping patches of width $w$, where $w=8$ for all streams in our experiments. Each timestep is augmented with a learnable positional embedding $q_t\in\mathbb{R}^{d_p}$ with $d_p=16$. For each patch, scalar values and positional codes are concatenated as follows:
\[
  u_i=[\tilde{s}_t;q_t]_{t=(i-1)w+1}^{iw}\in\mathbb{R}^{w(1+d_p)}
\]

Each patch representation is projected into the hidden space of the pretrained language model via a multilayer perceptron $f^{\text{ts}}$ consisting of 5 layers with hidden width 5120, matching the LLM embedding dimension $d$. The encoder outputs one embedding per patch,
\[
  z_i=f^{\text{ts}}(u_i)\in\mathbb{R}^d,\quad Z=\{z_1,\dots,z_N\}
\]
This design captures localized temporal patterns at the patch level while preserving absolute numerical semantics through reversible normalization.

\begin{figure}[tb]
  \centering
  \includegraphics[width=1\linewidth]{LENS_training.pdf}
  \caption{\textbf{LENS Architecture.} The model accepts multimodal inputs consisting of description text (e.g., "Heart rate..."), instruction text (e.g., "Summarize the user's current mental well-being?") and raw time-series sensor streams (e.g., heart rate). The text is processed by a pretrained LLM text embedder ($f^{\text{text}}$), while time-series data is encoded by a trainable patch-based encoder ($f^{\text{ts}}$). The resulting embeddings are concatenated into a unified sequence ($H$) and processed by the LLM backbone to generate a natural language response ($Y$).}
  \label{fig:lens_training}
\end{figure}


\subsubsection{Architecture}
Given an instruction text $X=\{x_1,\dots,x_M\}$ and $K$ time–series streams $\{S^{(k)}\}_{k=1}^{K}$,
we inject the normalization metadata described above into the text.
After this step, the prompt $\tilde{X}$ contains special placeholder tokens \texttt{<ts>} and \texttt{</ts>} to mark the position of each stream.
The text $\tilde{X}$ is then tokenized and passed through the pretrained LLM embedding layer $f^{\text{text}}$,
producing a sequence of text embeddings $E_{\mathrm{text}}=\{e_1,\dots,e_L\}\in\mathbb{R}^{L\times d}$ (Figure~\ref{fig:lens_training}).
In parallel, each time–series stream $S^{(k)}$ is processed by the encoder, yielding patch embeddings
$Z^{(k)}=\{z^{(k)}_1,\dots,z^{(k)}_{N_k}\}$ with $z^{(k)}_i\in\mathbb{R}^{d}$.
The multimodal embedding sequence is formed by concatenating the text embeddings and the patch embeddings at the positions referenced by the placeholders, i.e.,
$H = \operatorname{concat}\big(E_{\mathrm{text}},\,Z^{(1)},\dots,Z^{(K)}\big) \in\mathbb{R}^{L_{\mathrm{mm}}\times d}$, which interleaves natural–language and time–series representations in a single ordered context.
This unified sequence is then fed into the subsequent transformer blocks of the pretrained LLM, enabling multimodal reasoning over both textual instructions and temporal patterns.

\subsubsection{Training}
\textbf{Stage 1: Encoder Alignment.} The first stage aims to establish a stable alignment between temporal features and their textual descriptions. We use the alignment dataset from ChatTS \citep{xie_chatts_2025}, which takes a QA form and probes time–series understanding across different signal types and temporal behaviors, enabling the encoder to capture trends, correlations, and local pattern variations rather than raw numeric fluctuations. However, our final objective is to generate clinically coherent symptom narratives rather than attribute-only descriptions.
To prevent the encoder from over-specializing on low-level attribute queries, we interleave a small portion of narrative and general QA samples into the alignment corpus (8:1:1 with alignment data), giving the model early exposure to natural question forms and narrative structure and better preparing it for downstream symptom-focused generation.

\textbf{Stage 2: Supervised Fine-Tuning on Symptom Narratives.}
After encoder alignment, we train the model to perform symptom-centered question answering and narrative generation. We fine-tune LENS on the two EMA-derived QA datasets described in Section~\ref{sec:lens_narrative_synthesis}, which respectively provide item-level supervision for single-symptom interpretation and summary-level supervision for multi-symptom synthesis.
To prevent the model from overfitting to a single response style and to enhance its ability to follow structured outputs, we additionally include an instruction-following (IF) dataset constructed from predefined templates.
This dataset exposes the model to consistent answer formats, encouraging stable response organization under different prompts.
Because real-world time-series queries vary widely in temporal resolution, we further interleave an alignment-random subset in which sequence lengths are uniformly sampled between 64 and 1024, enabling the model to generalize across heterogeneous sampling rates and time spans.
During supervised fine-tuning, the four datasets are mixed using a 0.3:0.3:0.2:0.2 ratio (item-level QA : summary-level QA : IF : alignment-random), which balances symptom narrative grounding, structural instruction patterns, and robustness to sequence-length variability.

Given an instruction text $X=\{x_1,\dots,x_M\}$ and $K$ time–series streams $\{S^{(k)}\}_{k=1}^{K}$,
the model is trained in a standard supervised autoregressive manner.
For a target response $Y=\{y_1,\dots,y_U\}$, the objective is
\[
  \mathcal{L} = - \sum_{t=1}^{U} \log p_{\phi}\big(y_t \mid X,\{S^{(k)}\}_{k=1}^{K}, y_{1:t-1}\big)
\]
where $\phi$ denotes all trainable parameters.
We adopt full-parameter fine-tuning, updating the time-series encoder, pretrained LLM, and the text embedder. See Appendix~\S\ref{apd:implementation_details} for hardware and batching details.

\section{Experiments}

\subsection{Evaluation Settings}

\textbf{Baselines.}
To contextualize LENS’s performance, we compare it with baselines that follow common strategies for modeling time-series signals with LLMs. Prior work often encodes time-series values as raw text~\citep{kim2024healthllm, gruver2023large}, so we adopt the same principles and utilize the Qwen2.5-14B~\citep{qwen3} with few-shot prompting as the backbone; we refer to this as \textbf{TS-Text}. Another promising approach converts time-series data into visual representations to support downstream reasoning~\citep{yoon2024my,liu-etal-2025-picture}. Following this approach, we transform each signal into a plot and generate narratives using the Qwen-2.5VL-32B model in a few-shot setting. This baseline is denoted as \textbf{TS-Image}. Inference for all baseline models utilizes default configurations to ensure a controlled comparison, and we utilize participant-level splits (approx. 70:15:15) to prevent information leakage, ensuring that all data from a given individual appears in only one split (Appendix~\S\ref{apd:implementation_details}).

\textbf{Metrics.}
We evaluate model performance using three categories of metrics. \textbf{(1) Linguistic Metrics}: Generated narratives are assessed with ROUGE-1/2/L~\cite{lin-2004-rouge}, BLEU-4~\cite{papineni-etal-2002-bleu}, METEOR~\cite{banerjee-lavie-2005-meteor}, and BERTScore~\cite{zhang2020bertscore}, which measure lexical overlap, structural consistency, and semantic similarity with reference narratives. \textbf{(2) Symptom-grounded Evaluation}: To assess clinical alignment, we employ a structured LLM-as-a-judge protocol using \texttt{gpt-4.1-mini} (See prompt template in Appendix~\S\ref{apd:prompt_templates}. The judge is queried with temperature set to 0 to ensure deterministic and reproducible evaluations). For each of the PHQ-related symptom categories, the judge first outputs a JSON record containing \texttt{ref\_presence}, \texttt{ref\_severity}, \texttt{pred\_presence}, and \texttt{pred\_severity}, indicating whether the model omits or hallucinates symptom dimensions and whether predicted severity is faithful to the ordinal reference. \textbf{(3) Item-level QA Evaluation}: For single-question outputs, evaluation reduces to a record specifying \texttt{ref\_severity} and \texttt{pred\_severity}; symptom presence is implicit in the question itself, allowing severity correctness to be measured without aggregating across narrative spans. Detailed definitions of coverage, presence-aware severity alignment, and the weighting procedure used for ordinal scoring are provided in Appendix~\S\ref{sec:metric_appendix}.

\textbf{User study.}
To assess the characteristics of LENS-generated narratives, we recruited 13 mental health experts to evaluate LENS (14B; zero-shot) along with the two baselines: TS-Image (32B, few-shot) and TS-Text (14B, few-shot), across four rating dimensions. Each example is a narrative paired with a table of 12 ground-truth symptoms and assessed as follows: (1) Comprehensiveness: “Does the narrative mention the symptom?”; (2) Accuracy: “Does the narrative accurately describe the symptom severity (including synonyms)?”; (3) Clinical Utility: “Does the narrative provide clinically useful information that informs care?”; and (4) Language Cohesion: “Is the narrative coherent, easy to understand, and focused on depression and anxiety symptoms?” Comprehensiveness and accuracy were collected as yes/no responses and grouped into five categories. For comprehensiveness: 0–2 = Very Poor, 3–5 = Poor, 6–7 = Adequate, 8–11 = Good, 12 = Excellent. For accuracy: 0–2 = Major Discrepancy, 3–5 = Substantial Discrepancy, 6–7 = Partial Agreement, 8–11 = High Agreement, 12 = Complete Agreement. Clinical utility and language cohesion were rated on a 1–5 Likert scale. In total, we evaluated 117 narratives, with each expert rating 9 narratives (3 samples from each model). Additional information regarding the user study and expert background is provided in Appendix~\S\ref{apd:user_study}. To compare the models across the four domains, we first computed per-rater mean scores for each model by averaging over the three evaluated samples, and then performed paired t-tests across raters with Bonferroni correction~\citep{weisstein2004bonferroni}. Effect sizes are reported using Cohen’s $d_z$~\citep{goulet2018review}.

\begin{table}[]
  \centering 
  \resizebox{\columnwidth}{!}{
    \begin{tabular}{@{}lcccc@{}}
      \toprule
      \multirow{2}{*}{\textbf{Method}} & \multicolumn{2}{c}{\textbf{Linguistic Metrics}} & \multicolumn{2}{c}{\textbf{LLM-as-a-Judge Metrics}} \\ \cmidrule(l){2-3} \cmidrule(l){4-5}
      & \textbf{ROUGE-L }    & \textbf{BERTScore}    & \textbf{Coverage}        & \textbf{Presence Alignment}        \\ \cmidrule(r){1-1} \cmidrule(l){2-3} \cmidrule(l){4-5}

      \multicolumn{5}{c}{\textbf{Summary-level Evaluation}} \\
      \midrule
      LENS &\textbf{0.409} &\textbf{0.775} &\textbf{0.801} &\textbf{0.601} \\
      TS-Text &0.151 &0.630 &0.614 &0.372 \\
      TS-Image &0.373 &0.764 &0.740 &0.579 \\
      \midrule
      \multicolumn{5}{c}{\textbf{Item-level Evaluation}} \\
      \midrule
      LENS &\textbf{0.603} &\textbf{0.832} & -- &\textbf{0.732} \\
      TS-Text &0.142 &0.604 & -- &0.665 \\
      TS-Image &0.136 &0.602 & -- &0.687 \\

      \bottomrule
    \end{tabular}
  }
  \caption{\textbf{Results on Summary and Item level generation.} Additional metrics are provided in Appendix~\S\ref{apd:additional_results}.}
  \label{tab:results_main} 
\end{table}

\subsection{Results} \label{sec:results}

\textbf{LENS consistently outperforms baselines in generating comprehensive clinical narratives.} In the summary-level evaluation (Table~\ref{tab:results_main}), LENS achieves the highest scores across all linguistic metrics, recording a ROUGE-L of 0.409 and a BERTScore of 0.775. This represents a 9.6\% and 1.4\% improvement over the strongest baseline, TS-Image, which achieved 0.373 and 0.764, respectively. The significantly lower performance of TS-Text (ROUGE-L=0.151) highlights the difficulty of generating cohesive summaries from time-series text descriptions alone, whereas LENS effectively integrates multi-modal signals to produce structurally and semantically robust narratives.

\textbf{LENS demonstrates superior grounding in clinical symptoms, minimizing hallucinations.} From a clinical perspective, LENS achieves the highest alignment with ground-truth diagnoses. In summary-level generation, it attains a Symptom Coverage score of 0.801 and a Presence Alignment score of 0.601. This indicates that LENS is more capable of capturing the full spectrum of patient symptoms ($K=14$ categories) without omitting critical indicators or fabricating non-existent ones, outperforming TS-Image (Coverage=0.740) and greatly surpassing TS-Text (Presence Alignment=0.372).

\textbf{LENS exhibits strong performance in fine-grained, item-level query answering.} Compared to summary-level narrative generation, the performance gap in the item-level evaluation is more pronounced, indicating that the task requires more precise retrieval (Table \ref{tab:results_main}). For example, LENS achieves a ROUGE-L score of 0.603, which is more than four times higher than both TS-Text (0.142) and TS-Image (0.136). This trend holds for semantic and clinical metrics as well, with LENS securing the highest Presence Alignment of 0.732 (10.1\% improvement over TS-Text and 6.6\% over TS-Image) and a BERTScore of 0.832 (over 37\% improvement against both baselines). These results suggest that while baselines struggle to isolate specific symptom intensities in a QA format, LENS maintains high fidelity to the reference severity levels.

\begin{figure}
  \centering
  \includegraphics[width=1\linewidth]{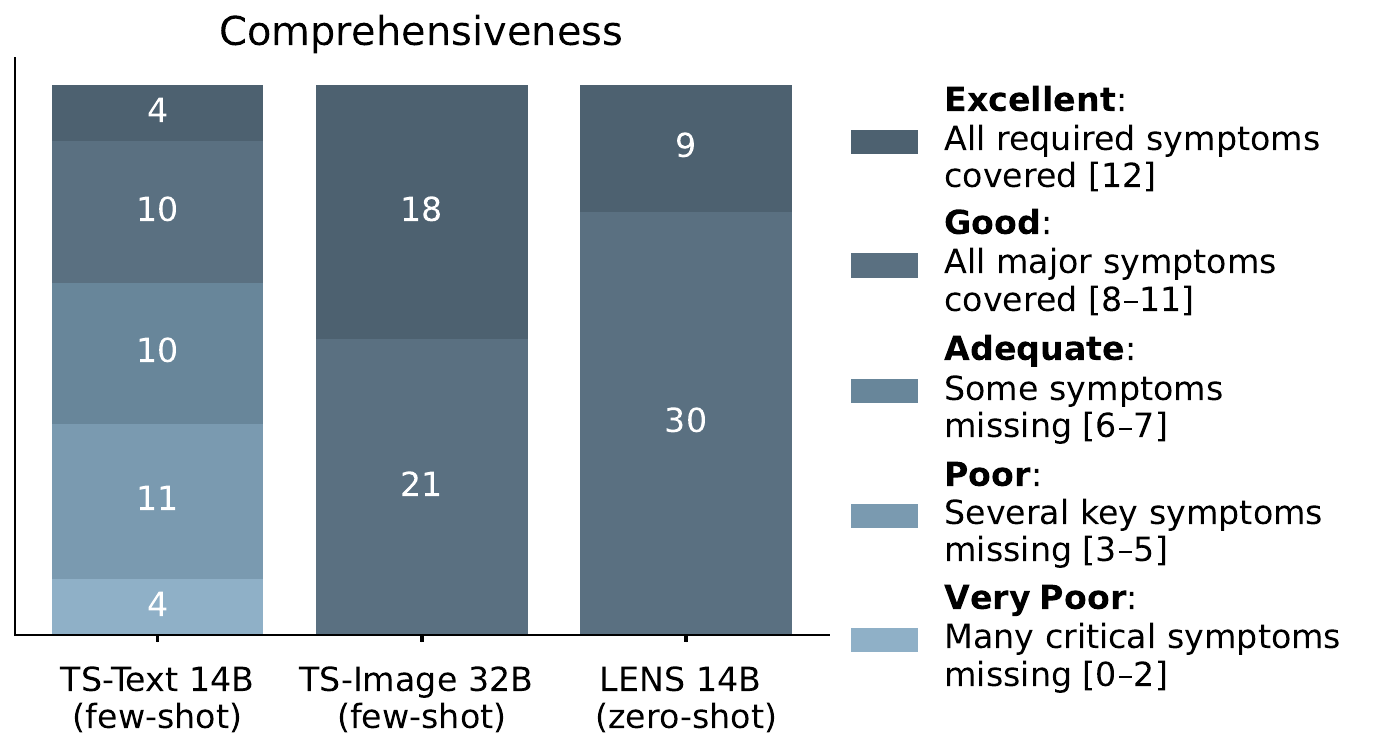}
  \caption{\textbf{User Study: Comprehensiveness.} Expert ratings compare how many symptoms each model’s narrative successfully covers.}
  \label{fig:user_study_comprehensiveness}
\end{figure}

\begin{figure}
  \centering
  \includegraphics[width=\linewidth]{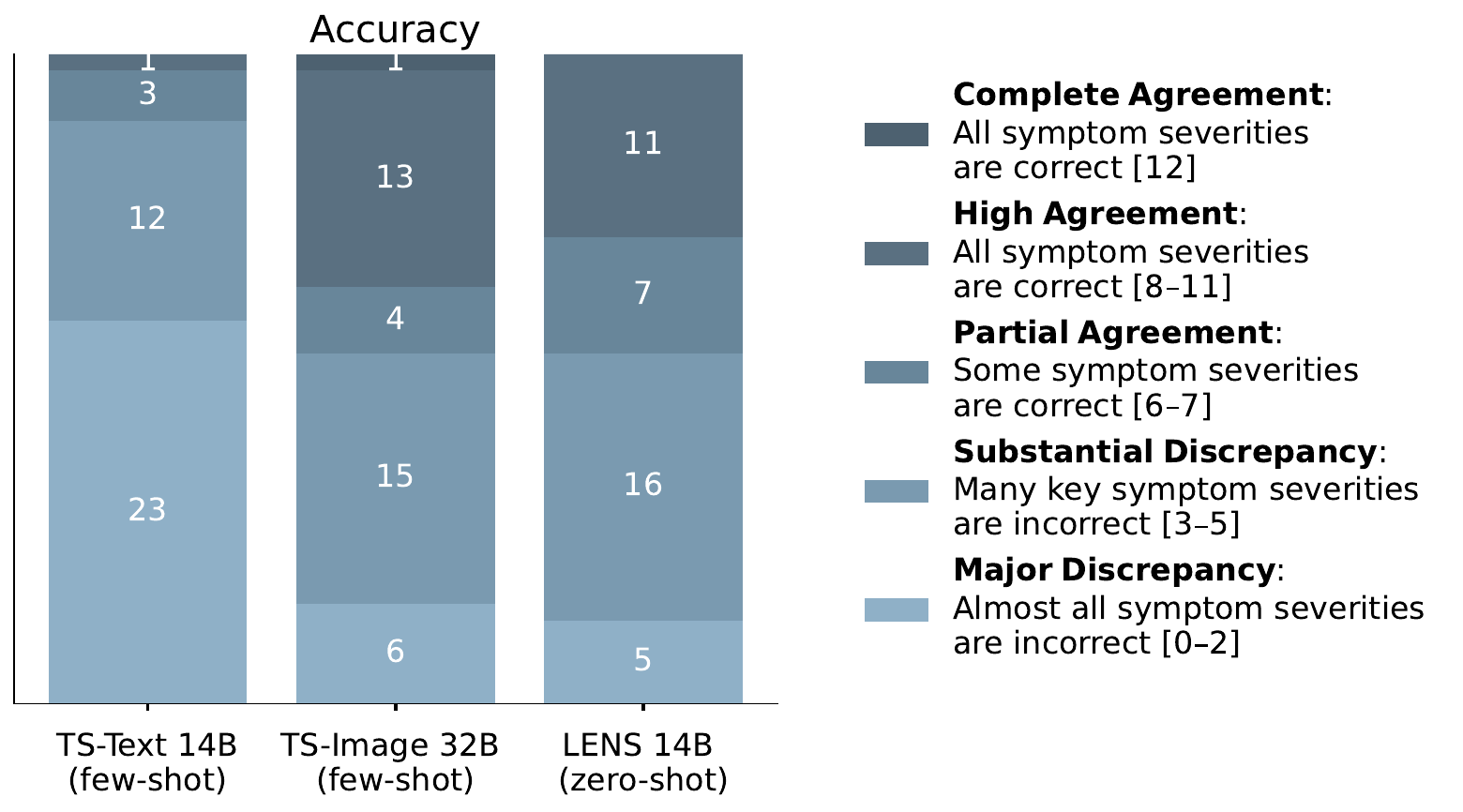}
  \caption{\textbf{User Study: Accuracy.} Expert ratings assess how well each model’s narrative captures the severity of symptoms reported in the ground-truth EMA.}   \label{fig:user_study_accuracy}
\end{figure}

\begin{figure}
  \centering
  \includegraphics[width=\linewidth]{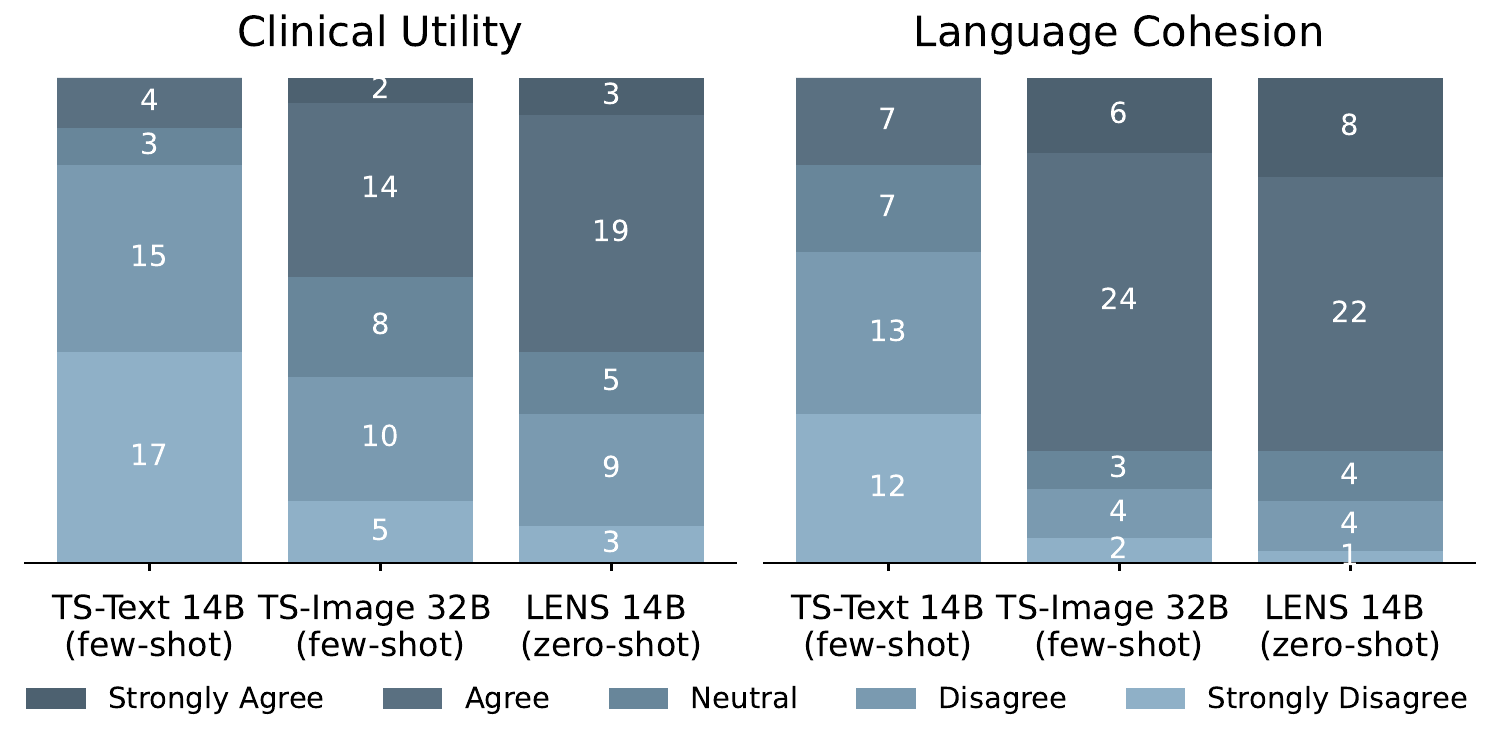}
  \caption{\textbf{User Study: Clinical Utility \& Language Cohesion.} Comparing expert ratings for the usefulness of the narratives and the cohesiveness of the language.}
  \label{fig:user_study_uc}
\end{figure}

\textbf{Mental health experts rate LENS narratives significantly higher than TS-Text.} As shown in Figures~\ref{fig:user_study_comprehensiveness} to \ref{fig:user_study_uc}, LENS consistently outperforms TS-Text across all four evaluation dimensions. Specifically, LENS narratives receive significantly higher expert ratings for comprehensiveness (4.23 vs. 2.97; $T = 7.02$; $p < 0.01$; $d_z = 1.12$), accuracy (2.61 vs. 1.53; $T = 6.49$; $p < 0.01$; $d_z = 1.04$), clinical utility (3.25 vs. 1.84; $T = 7.16$; $p < 0.01$; $d_z = 1.15$), and language cohesion (3.82 vs. 2.23; $T = 7.34$; $p < 0.01$; $d_z = 1.17$). These results demonstrate that experts find narratives derived from native time-series integration to be not only more accurate but also more practically useful for care than those generated from text descriptions alone.

\textbf{Expert evaluation demonstrates performance parity with larger visual baselines.} Comparing LENS to TS-Image (Figures~\ref{fig:user_study_comprehensiveness} to \ref{fig:user_study_uc}), we observe no statistically significant differences between the two models across any of the four evaluation dimensions. Expert ratings are comparable for comprehensiveness (4.23 vs. 4.46; $T=-2.47$; $p > 0.01$; $d_z = -0.39$), accuracy (2.61 vs. 2.69; $T=-0.42$; $p > 0.01$; $d_z=-0.06$), clinical utility (3.82 vs. 3.71; $T=1.52$; $p > 0.01$; $d_z=0.24$), and language cohesion (3.25 vs. 2.94; $T=0.56$; $p > 0.01$; $d_z=0.09$). These results indicate that LENS matches the performance of the $2.2\times$ larger TS-Image while offering a significantly streamlined inference process. By replacing complex visual plot engineering with native time-series encoding, LENS achieves these results in a zero-shot setting. Interestingly, experts identified LENS as having better real-world applicability, evidenced by its marginally higher ratings for clinical utility and language cohesion.

\textbf{Ablation and scalability analysis.} We analyze LENS' architecture through ablation and scaling studies detailed in Appendix~\S\ref{apd:additional_results}. An evaluation comparing LENS to a fine-tuned text-only baseline (TS-Text-FT) reveals that explicit time-series encoding is essential for capturing temporal structures; LENS significantly outperforms the text-only variant, particularly in item-level QA where the ROUGE-L gap is markedly wider (0.6030 vs. 0.1717) (Table \ref{tab:ablation_textft}). Furthermore, LENS demonstrates superior computational efficiency, requiring approximately 930 tokens per sample, a reduction of roughly 94\% compared to verbose text serialization and 4$\times$ relative to vision-based models (Figure \ref{fig:token_consumption}). We also observe that while basic instruction-following can be established with 10\% of the training data, the full dataset is necessary to maximize clinical fidelity in complex narratives, with Presence Alignment improving as data size increases (Table \ref{tab:data_efficiency}). Ablations on encoder hyperparameters further confirm robustness: varying MLP depth ($L\in\{3,5,7\}$) and patch size ($p\in\{4,8,16\}$) yields fluctuations within 0.01 absolute across all metrics (Table~\ref{tab:ablation_encoder}). Finally, experiments with a lighter LENS-7B variant show that it maintains linguistic performance (ROUGE-L 0.409 vs. 0.410) while trailing slightly in complex clinical alignment (0.559 vs. 0.601) (Table \ref{tab:model_size_comparison}). Additionally, a sensor mismatch experiment in which sensor streams are randomly shuffled across participants reveals that Presence Alignment drops by 20.1\% while linguistic metrics remain unchanged (Table~\ref{tab:sensor_mismatch}), providing direct evidence that LENS grounds clinical content in raw sensor signals rather than merely reproducing stylistic patterns from the training labels.

\section{Conclusion}
In this work, we introduced LENS, a framework for generating clinical narratives by aligning raw sensor streams with LLMs. Through a scalable data-synthesis pipeline and a patch-based time-series encoder, LENS addresses two key challenges: the scarcity of paired sensor–text data and the difficulty of processing long-horizon numerical signals within LLMs. Using a dataset of 50,957 samples, our evaluation showed that LENS produced linguistically fluent and clinically accurate descriptions of depression and anxiety symptoms. LENS also matched the performance of $2.2\times$ larger vision-based models while avoiding the overhead of plot generation, and mental-health experts rated its narratives as more applicable and clinically useful than text-based baselines. Future work should examine LENS’s transferability to broader demographics and additional psychological instruments. Overall, LENS represents a fundamental step toward ecologically valid mental-health monitoring by transforming raw sensor data into interpretable clinical insights.

\section*{Limitations}
Several limitations of this work warrant consideration. First, LENS currently focuses on depression and anxiety within a specific clinical population. While these conditions are well-suited for establishing a sensor-to-language framework, the findings may not yet generalize to other demographics, psychological instruments, or broader populations. Given the pipeline’s modular design, future work should evaluate its extensibility to alternative disorders. Second, the framework analyzes behavior over four-hour windows. This captures immediate states but may miss longitudinal trends, such as relapse signals or treatment responses that unfold over months. Future extensions could employ hierarchical modeling or memory-augmented architectures to integrate these short-term representations into longer-horizon reasoning.

Third, this study utilized models from the Qwen family to maintain a controlled experimental environment rather than to optimize for peak performance. Consequently, we did not assess cross-model robustness. Future research should examine how different architectures, including instruction-tuned or domain-adapted variants, vary in their ability to interpret sensor data. Fourth, passive sensing lacks the situational context necessary to fully disambiguate certain signals. For instance, an elevated heart rate could indicate physical exertion, illness, or acute stress. Incorporating richer contextual data or structured user input remains a necessary step for reducing such ambiguity. Finally, the iterative Refine-n-Judge loop improves output quality but introduces computational overhead. This may challenge real-time deployment or large-scale application under constrained budgets. Exploring model distillation or more efficient training strategies could help mitigate these scaling issues.

\section*{Ethical Considerations}
\textbf{Study Participation.} IRB-approved procedures involved 258 analyzed participants (from 300 recruited) who provided informed consent. Inclusion required a SCID-verified MDD diagnosis, excluding bipolar disorder, psychosis, or active suicidality. The sample was 84\% female and 79\% White (12\% Hispanic/Latino) with a mean age of 40. Most participants were college-educated (93\%) and employed (61\%), with incomes aligning with national distributions. Compensation included \$1 per completed EMA and a \$50 bonus for 90\% completion. Safety was maintained via automated suicidality alerts and continuous clinical oversight.

\textbf{Systems \& Software.}
Modeling was conducted using local LLMs within a secure, closed environment to prevent data transmission to external APIs or cloud services. Access was restricted to authorized, IRB-trained researchers via role-based controls and audit logging. Collectively, these safeguards reduced the risk of re-identification of personally identifiable information and ensured adherence to IRB-approved data protection and privacy protocols.

\textbf{Adverse Usage.} While LLMs offer significant potential for interpreting mental health time-series data, their application requires caution to protect participant safety and clinical integrity. These narratives must augment existing clinical workflows rather than serve as standalone diagnostic tools, as over-reliance on automated summaries may obscure critical nuances within the raw sensor data. Furthermore, the utilization of sensitive signals, such as GPS traces, necessitates strict governance to prevent unauthorized surveillance or secondary data use. Given that LLMs cannot entirely eliminate the risk of hallucinations, all generated outputs must be reviewed by qualified experts to ensure clinical judgment remains the final authority for interpretation and intervention.

\section*{Acknowledgments}
The research presented in this paper was supported by the \textbf{National Institute of Mental Health (NIMH)} under award number \textbf{R01MH123482-01}, and by \textbf{Evergreen: A Generative AI and Behavioral Sensing Digital Ecosystem to Promote Student Wellness and Flourishing.} This work was further enabled by philanthropic gifts to Dartmouth College dedicated to advancing AI-supported well-being and student flourishing. The contents of this manuscript are solely the responsibility of the authors and do not necessarily represent the views of the funding agencies. The funders had no role in the study design, data collection, analysis, interpretation, or the preparation of this manuscript.

\bibliography{custom}

@article{kroenke2001phq,
  title={The PHQ-9: validity of a brief depression severity measure},
  author={Kroenke, Kurt and Spitzer, Robert L and Williams, Janet BW},
  journal={Journal of general internal medicine},
  volume={16},
  number={9},
  pages={606--613},
  year={2001},
  publisher={Wiley Online Library}
}

@article{abd2023systematic,
  title={Systematic review and meta-analysis of performance of wearable artificial intelligence in detecting and predicting depression},
  author={Abd-Alrazaq, Alaa and AlSaad, Rawan and Shuweihdi, Farag and Ahmed, Arfan and Aziz, Sarah and Sheikh, Javaid},
  journal={NPJ Digital Medicine},
  volume={6},
  number={1},
  pages={84},
  year={2023},
  publisher={Nature Publishing Group UK London}
}

@article{nepal2024social,
  title={Social isolation and serious mental illness: the role of context-aware mobile interventions},
  author={Nepal, Subigya and Pillai, Arvind and Parrish, Emma M and Holden, Jason and Depp, Colin and Campbell, Andrew T and Granholm, Eric L},
  journal={IEEE pervasive computing},
  volume={23},
  number={1},
  pages={46--56},
  year={2024},
  publisher={IEEE}
}

@article{goulet2018review,
  title={A review of effect sizes and their confidence intervals, Part I: The Cohen’sd family},
  author={Goulet-Pelletier, Jean-Christophe and Cousineau, Denis},
  journal={The Quantitative Methods for Psychology},
  volume={14},
  number={4},
  pages={242--265},
  year={2018},
  publisher={The Quantitative Methods for Psychology}
}

@article{weisstein2004bonferroni,
  title={Bonferroni correction},
  author={Weisstein, Eric W},
  journal={https://mathworld. wolfram. com/},
  year={2004},
  publisher={Wolfram Research, Inc.}
}

@inproceedings{pillai2023rare,
  title={Rare life event detection via mobile sensing using multi-task learning},
  author={Pillai, Arvind and Nepal, Subigya and Campbell, Andrew},
  booktitle={Conference on Health, Inference, and Learning},
  pages={279--293},
  year={2023},
  organization={PMLR}
}

@misc{qwen3,
    title={Qwen3 Technical Report}, 
    author={An Yang and Anfeng Li and Baosong Yang and Beichen Zhang and Binyuan Hui and Bo Zheng and Bowen Yu and Chang Gao and Chengen Huang and Chenxu Lv and Chujie Zheng and Dayiheng Liu and Fan Zhou and Fei Huang and Feng Hu and Hao Ge and Haoran Wei and Huan Lin and Jialong Tang and Jian Yang and Jianhong Tu and Jianwei Zhang and Jianxin Yang and Jiaxi Yang and Jing Zhou and Jingren Zhou and Junyang Lin and Kai Dang and Keqin Bao and Kexin Yang and Le Yu and Lianghao Deng and Mei Li and Mingfeng Xue and Mingze Li and Pei Zhang and Peng Wang and Qin Zhu and Rui Men and Ruize Gao and Shixuan Liu and Shuang Luo and Tianhao Li and Tianyi Tang and Wenbiao Yin and Xingzhang Ren and Xinyu Wang and Xinyu Zhang and Xuancheng Ren and Yang Fan and Yang Su and Yichang Zhang and Yinger Zhang and Yu Wan and Yuqiong Liu and Zekun Wang and Zeyu Cui and Zhenru Zhang and Zhipeng Zhou and Zihan Qiu},
    journal = {arXiv preprint arXiv:2505.09388},
    year={2025}
}

@misc{cayir2025refinenjudge,
      title={Refine-n-Judge: Curating High-Quality Preference Chains for LLM-Fine-Tuning}, 
      author={Derin Cayir and Renjie Tao and Rashi Rungta and Kai Sun and Sean Chen and Haidar Khan and Minseok Kim and Julia Reinspach and Yue Liu},
      year={2025},
      eprint={2508.01543},
      archivePrefix={arXiv},
      primaryClass={cs.AI},
      url={https://arxiv.org/abs/2508.01543}, 
}

@article{choudhary2022machineML,
  title={A machine learning approach for detecting digital behavioral patterns of depression using nonintrusive smartphone data (complementary path to patient health questionnaire-9 assessment): Prospective observational study},
  author={Choudhary, Soumya and Thomas, Nikita and Ellenberger, Janine and Srinivasan, Girish and Cohen, Roy},
  journal={JMIR Formative Research},
  volume={6},
  number={5},
  pages={e37736},
  year={2022},
  publisher={JMIR Publications Inc., Toronto, Canada}
}

@inproceedings{YangMentaLLaMA2024,
author = {Yang, Kailai and Zhang, Tianlin and Kuang, Ziyan and Xie, Qianqian and Huang, Jimin and Ananiadou, Sophia},
title = {MentaLLaMA: Interpretable Mental Health Analysis on Social Media with Large Language Models},
year = {2024},
isbn = {9798400701719},
publisher = {Association for Computing Machinery},
address = {New York, NY, USA},
url = {https://doi-org.dartmouth.idm.oclc.org/10.1145/3589334.3648137},
doi = {10.1145/3589334.3648137},
booktitle = {Proceedings of the ACM Web Conference 2024},
pages = {4489–4500},
numpages = {12},
location = {Singapore, Singapore},
series = {WWW '24}
}

@misc{moon2025depressllm,
  title={DepressLLM: Interpretable domain-adapted language model for depression detection from real-world narratives},
  author={Moon, Sehwan and Lee, Aram and Kim, Jeong Eun and Kang, Hee-Ju and Shin, Il-Seon and Kim, Sung-Wan and Kim, Jae-Min and Jhon, Min and Kim, Ju-Wan},
  journal={arXiv preprint arXiv:2508.08591},
  year={2025}
}

@misc{kim2024healthllm,
  title={Health-llm: Large language models for health prediction via wearable sensor data},
  author={Kim, Yubin and Xu, Xuhai and McDuff, Daniel and Breazeal, Cynthia and Park, Hae Won},
  journal={arXiv preprint arXiv:2401.06866},
  year={2024}
}

@misc{li2022blip,
      title={BLIP: Bootstrapping Language-Image Pre-training for Unified Vision-Language Understanding and Generation}, 
      author={Junnan Li and Dongxu Li and Caiming Xiong and Steven Hoi},
      year={2022},
      eprint={2201.12086},
      archivePrefix={arXiv},
      primaryClass={cs.CV},
      url={https://arxiv.org/abs/2201.12086}, 
}

@misc{mistral7b,
      title={Mistral 7B}, 
      author={Albert Q. Jiang and Alexandre Sablayrolles and Arthur Mensch and Chris Bamford and Devendra Singh Chaplot and Diego de las Casas and Florian Bressand and Gianna Lengyel and Guillaume Lample and Lucile Saulnier and L\'{e}lio Renard Lavaud and Marie-Anne Lachaux and Pierre Stock and Teven Le Scao and Thibaut Lavril and Thomas Wang and Timoth\'{e}e Lacroix and William El Sayed},
      year={2023},
      eprint={2310.06825},
      archivePrefix={arXiv},
      primaryClass={cs.CL},
      url={https://arxiv.org/abs/2310.06825}, 
}

@misc{qwen2025qwen25technicalreport,
      title={Qwen2.5 Technical Report}, 
      author={Qwen and : and An Yang and Baosong Yang and Beichen Zhang and Binyuan Hui and Bo Zheng and Bowen Yu and Chengyuan Li and Dayiheng Liu and Fei Huang and Haoran Wei and Huan Lin and Jian Yang and Jianhong Tu and Jianwei Zhang and Jianxin Yang and Jiaxi Yang and Jingren Zhou and Junyang Lin and Kai Dang and Keming Lu and Keqin Bao and Kexin Yang and Le Yu and Mei Li and Mingfeng Xue and Pei Zhang and Qin Zhu and Rui Men and Runji Lin and Tianhao Li and Tianyi Tang and Tingyu Xia and Xingzhang Ren and Xuancheng Ren and Yang Fan and Yang Su and Yichang Zhang and Yu Wan and Yuqiong Liu and Zeyu Cui and Zhenru Zhang and Zihan Qiu},
      year={2025},
      eprint={2412.15115},
      archivePrefix={arXiv},
      primaryClass={cs.CL},
      url={https://arxiv.org/abs/2412.15115}, 
}

@inproceedings{liu-etal-2025-picture,
    title = "A Picture is Worth A Thousand Numbers: Enabling {LLM}s Reason about Time Series via Visualization",
    author = "Liu, Haoxin  and
      Liu, Chenghao  and
      Prakash, B. Aditya",
    editor = "Chiruzzo, Luis  and
      Ritter, Alan  and
      Wang, Lu",
    booktitle = "Proceedings of the 2025 Conference of the Nations of the Americas Chapter of the Association for Computational Linguistics: Human Language Technologies (Volume 1: Long Papers)",
    month = apr,
    year = "2025",
    address = "Albuquerque, New Mexico",
    publisher = "Association for Computational Linguistics",
    url = "https://aclanthology.org/2025.naacl-long.383/",
    doi = "10.18653/v1/2025.naacl-long.383",
    pages = "7486--7518",
    ISBN = "979-8-89176-189-6"
}

@misc{llama3,
      title={The Llama 3 Herd of Models}, 
      author={Aaron Grattafiori and Abhimanyu Dubey and Abhinav Jauhri and others},
      year={2024},
      eprint={2407.21783},
      archivePrefix={arXiv},
      primaryClass={cs.AI},
      url={https://arxiv.org/abs/2407.21783}, 
}

@inproceedings{papineni-etal-2002-bleu,
    title = "{B}leu: a Method for Automatic Evaluation of Machine Translation",
    author = "Papineni, Kishore  and
      Roukos, Salim  and
      Ward, Todd  and
      Zhu, Wei-Jing",
    editor = "Isabelle, Pierre  and
      Charniak, Eugene  and
      Lin, Dekang",
    booktitle = "Proceedings of the 40th Annual Meeting of the Association for Computational Linguistics",
    month = jul,
    year = "2002",
    address = "Philadelphia, Pennsylvania, USA",
    publisher = "Association for Computational Linguistics",
    url = "https://aclanthology.org/P02-1040/",
    doi = "10.3115/1073083.1073135",
    pages = "311--318"
}

@misc{zhang2020bertscore,
      title={BERTScore: Evaluating Text Generation with BERT}, 
      author={Tianyi Zhang and Varsha Kishore and Felix Wu and Kilian Q. Weinberger and Yoav Artzi},
      year={2020},
      eprint={1904.09675},
      archivePrefix={arXiv},
      primaryClass={cs.CL},
      url={https://arxiv.org/abs/1904.09675}, 
}

@article{GLOBEM,
author = {Xu, Xuhai and Liu, Xin and Zhang, Han and Wang, Weichen and Nepal, Subigya and others},
title = {GLOBEM: Cross-Dataset Generalization of Longitudinal Human Behavior Modeling},
year = {2023},
issue_date = {December 2022},
publisher = {Association for Computing Machinery},
address = {New York, NY, USA},
volume = {6},
number = {4},
url = {https://doi-org.dartmouth.idm.oclc.org/10.1145/3569485},
doi = {10.1145/3569485},
journal = {Proc. ACM Interact. Mob. Wearable Ubiquitous Technol.},
month = jan,
articleno = {190},
numpages = {34}
}

@inproceedings{lin-2004-rouge,
    title = "{ROUGE}: A Package for Automatic Evaluation of Summaries",
    author = "Lin, Chin-Yew",
    booktitle = "Text Summarization Branches Out",
    month = jul,
    year = "2004",
    address = "Barcelona, Spain",
    publisher = "Association for Computational Linguistics",
    url = "https://aclanthology.org/W04-1013/",
    pages = "74--81"
}

@inproceedings{banerjee-lavie-2005-meteor,
    title = "{METEOR}: An Automatic Metric for {MT} Evaluation with Improved Correlation with Human Judgments",
    author = "Banerjee, Satanjeev  and
      Lavie, Alon",
    editor = "Goldstein, Jade  and
      Lavie, Alon  and
      Lin, Chin-Yew  and
      Voss, Clare",
    booktitle = "Proceedings of the {ACL} Workshop on Intrinsic and Extrinsic Evaluation Measures for Machine Translation and/or Summarization",
    month = jun,
    year = "2005",
    address = "Ann Arbor, Michigan",
    publisher = "Association for Computational Linguistics",
    url = "https://aclanthology.org/W05-0909/",
    pages = "65--72"
}

@misc{radford2021Clip,
      title={Learning Transferable Visual Models From Natural Language Supervision}, 
      author={Alec Radford and Jong Wook Kim and Chris Hallacy and Aditya Ramesh and Gabriel Goh and Sandhini Agarwal and Girish Sastry and Amanda Askell and Pamela Mishkin and Jack Clark and Gretchen Krueger and Ilya Sutskever},
      year={2021},
      eprint={2103.00020},
      archivePrefix={arXiv},
      primaryClass={cs.CV},
      url={https://arxiv.org/abs/2103.00020}, 
}

@misc{dosovitskiy2021ViT,
      title={An Image is Worth 16x16 Words: Transformers for Image Recognition at Scale}, 
      author={Alexey Dosovitskiy and Lucas Beyer and Alexander Kolesnikov and Dirk Weissenborn and Xiaohua Zhai and Thomas Unterthiner and Mostafa Dehghani and Matthias Minderer and Georg Heigold and Sylvain Gelly and Jakob Uszkoreit and Neil Houlsby},
      year={2021},
      eprint={2010.11929},
      archivePrefix={arXiv},
      primaryClass={cs.CV},
      url={https://arxiv.org/abs/2010.11929}, 
}

@inproceedings{tan2024language,
author = {Tan, Mingtian and Merrill, Mike A. and Gupta, Vinayak and Althoff, Tim and Hartvigsen, Thomas},
title = {Are language models actually useful for time series forecasting?},
year = {2025},
isbn = {9798331314385},
publisher = {Curran Associates Inc.},
address = {Red Hook, NY, USA},
booktitle = {Proceedings of the 38th International Conference on Neural Information Processing Systems},
articleno = {1922},
numpages = {30},
location = {Vancouver, BC, Canada},
series = {NIPS '24}
}

@misc{li_sensorllm_2025,
    title = {{SensorLLM}: {Human}-{Intuitive} {Alignment} of {Multivariate} {Sensor} {Data} with {LLMs} for {Activity} {Recognition}},
    shorttitle = {{SensorLLM}},
    url = {http://arxiv.org/abs/2410.10624},
    doi = {10.48550/arXiv.2410.10624},
    language = {en},
    urldate = {2025-06-18},
    publisher = {arXiv},
    author = {Li, Zechen and Deldari, Shohreh and Chen, Linyao and Xue, Hao and Salim, Flora D.},
    month = may,
    year = {2025},
    note = {arXiv:2410.10624 [cs]},
}

@misc{xie_chatts_2025,
    title = {{ChatTS}: {Aligning} {Time} {Series} with {LLMs} via {Synthetic} {Data} for {Enhanced} {Understanding} and {Reasoning}},
    shorttitle = {{ChatTS}},
    url = {http://arxiv.org/abs/2412.03104},
    doi = {10.48550/arXiv.2412.03104},
    urldate = {2025-08-26},
    publisher = {arXiv},
    author = {Xie, Zhe and Li, Zeyan and He, Xiao and Xu, Longlong and Wen, Xidao and Zhang, Tieying and Chen, Jianjun and Shi, Rui and Pei, Dan},
    month = apr,
    year = {2025},
    note = {arXiv:2412.03104 [cs]},
}

@misc{opentslm,
      title={OpenTSLM: Time-Series Language Models for Reasoning over Multivariate Medical Text- and Time-Series Data}, 
      author={Patrick Langer and Thomas Kaar and Max Rosenblattl and Maxwell A. Xu and Winnie Chow and Martin Maritsch and Aradhana Verma and Brian Han and Daniel Seung Kim and Henry Chubb and Scott Ceresnak and Aydin Zahedivash and Alexander Tarlochan Singh Sandhu and Fatima Rodriguez and Daniel McDuff and Elgar Fleisch and Oliver Aalami and Filipe Barata and Paul Schmiedmayer},
      year={2025},
      eprint={2510.02410},
      archivePrefix={arXiv},
      primaryClass={cs.LG},
      url={https://arxiv.org/abs/2510.02410}, 
}

@misc{yan_voila-_nodate,
      title={Voila-A: Aligning Vision-Language Models with User's Gaze Attention}, 
      author={Kun Yan and Lei Ji and Zeyu Wang and Yuntao Wang and Nan Duan and Shuai Ma},
      year={2023},
      eprint={2401.09454},
      archivePrefix={arXiv},
      primaryClass={cs.CV},
      url={https://arxiv.org/abs/2401.09454}, 
}

@misc{imran_llasa_2024,
    title = {{LLaSA}: {A} {Multimodal} {LLM} for {Human} {Activity} {Analysis} {Through} {Wearable} and {Smartphone} {Sensors}},
    shorttitle = {{LLaSA}},
    url = {http://arxiv.org/abs/2406.14498},
    doi = {10.48550/arXiv.2406.14498},
    language = {en-US},
    urldate = {2025-05-05},
    publisher = {arXiv},
    author = {Imran, Sheikh Asif and Khan, Mohammad Nur Hossain and Biswas, Subrata and Islam, Bashima},
    month = dec,
    year = {2024},
    note = {arXiv:2406.14498 [cs]
version: 2},
}

@inproceedings{wang2014studentlife,
  title={StudentLife: assessing mental health, academic performance and behavioral trends of college students using smartphones},
  author={Wang, Rui and Chen, Fanglin and Chen, Zhenyu and Li, Tianxing and Harari, Gabriella and Tignor, Stefanie and Zhou, Xia and Ben-Zeev, Dror and Campbell, Andrew T},
  booktitle={Proceedings of the 2014 ACM international joint conference on pervasive and ubiquitous computing},
  pages={3--14},
  year={2014}
}

@misc{justin2024towards,
    title = {Towards a personal health large language model},
    url = {https://www.arxiv.org/abs/2406.06474},
    journal = {arXiv preprint arXiv:2406.06474},
    author = {Justin, Cosentino and Anastasiya, Belyaeva and Xin, Liu and Nicholas, A., Furlotte and Zhun, Yang and Chace, Lee and Erik, Schenck and Yojan, Patel and Jian, Cui and Logan, Douglas, Schneider and Robby, Bryant and Ryan, G., Gomes and Allen, Jiang and Roy, Lee and Yun, Liu and Javier, Perez and Jameson, K., Rogers and Cathy, Speed and Shyam, Tailor and Megan, Walker and Jeffrey, Yu and Tim, Althoff and Conor, Heneghan and John, Hernandez and Mark, Malhotra and Leor, Stern and Yossi, Matias and Greg, S., Corrado and Shwetak, Patel and Shravya, Shetty and Jiening, Zhan and Shruthi, Prabhakara and Daniel, McDuff and Cory, Y., McLean},
    year = {2024},
}

@article{englhardt_classification_2024,
    title = {From {Classification} to {Clinical} {Insights}: {Towards} {Analyzing} and {Reasoning} {About} {Mobile} and {Behavioral} {Health} {Data} {With} {Large} {Language} {Models}},
    volume = {8},
    issn = {2474-9567},
    shorttitle = {From {Classification} to {Clinical} {Insights}},
    url = {http://arxiv.org/abs/2311.13063},
    doi = {10.1145/3659604},
    language = {en-US},
    number = {2},
    urldate = {2025-04-23},
    journal = {Proceedings of the ACM on Interactive, Mobile, Wearable and Ubiquitous Technologies},
    author = {Englhardt, Zachary and Ma, Chengqian and Morris, Margaret E. and Xu, Xuhai "Orson" and Chang, Chun-Cheng and Qin, Lianhui and McDuff, Daniel and Liu, Xin and Patel, Shwetak and Iyer, Vikram},
    month = may,
    year = {2024},
    note = {arXiv:2311.13063 [cs]},
    pages = {1--25},
}

@misc{liu2023visualinstructiontuning,
      title={Visual Instruction Tuning}, 
      author={Haotian Liu and Chunyuan Li and Qingyang Wu and Yong Jae Lee},
      year={2023},
      eprint={2304.08485},
      archivePrefix={arXiv},
      primaryClass={cs.CV},
      url={https://arxiv.org/abs/2304.08485}, 
}

@article{xue2023promptcast,
  title={Promptcast: A new prompt-based learning paradigm for time series forecasting},
  author={Xue, Hao and Salim, Flora D},
  journal={IEEE Transactions on Knowledge and Data Engineering},
  volume={36},
  number={11},
  pages={6851--6864},
  year={2023},
  publisher={IEEE}
}

@article{spathis2024first,
  title={The first step is the hardest: Pitfalls of representing and tokenizing temporal data for large language models},
  author={Spathis, Dimitris and Kawsar, Fahim},
  journal={Journal of the American Medical Informatics Association},
  volume={31},
  number={9},
  pages={2151--2158},
  year={2024},
  publisher={Oxford Academic}
}

@misc{
zhang2023insight,
title={Insight Miner: A Large-scale Multimodal Model for Insight Mining from Time Series},
author={Yunkai Zhang and Yawen Zhang and Ming Zheng and Kezhen Chen and Chongyang Gao and Ruian Ge and Siyuan Teng and Amine Jelloul and Jinmeng Rao and Xiaoyuan Guo and Chiang-Wei Fang and Zeyu Zheng and Jie Yang},
booktitle={NeurIPS 2023 AI for Science Workshop},
year={2023},
url={https://openreview.net/forum?id=E1khscdUdH}
}

@misc{pillai_beyond_2025,
    title = {Beyond {Prompting}: {Time2Lang} -- {Bridging} {Time}-{Series} {Foundation} {Models} and {Large} {Language} {Models} for {Health} {Sensing}},
    shorttitle = {Beyond {Prompting}},
    url = {http://arxiv.org/abs/2502.07608},
    doi = {10.48550/arXiv.2502.07608},
    urldate = {2025-02-23},
    publisher = {arXiv},
    author = {Pillai, Arvind and Spathis, Dimitris and Nepal, Subigya and Collins, Amanda C. and Mackin, Daniel M. and Heinz, Michael V. and Griffin, Tess Z. and Jacobson, Nicholas C. and Campbell, Andrew},
    month = feb,
    year = {2025},
    note = {arXiv:2502.07608 [cs]},
}

@misc{ming2023timellm,
    title = {Time-{LLM}: {Time} series forecasting by reprogramming large language models},
    url = {https://www.arxiv.org/abs/2310.01728},
    journal = {arXiv preprint arXiv:2310.01728},
    author = {Ming, Jin and Shiyu, Wang and Lintao, Ma and Zhixuan, Chu and James, Y., Zhang and Xiaoming, Shi and Pin-Yu, Chen and Yuxuan, Liang and Yuan-Fang, Li and Shirui, Pan and Qingsong, Wen},
    year = {2023},
}

@misc{li_generation_2025,
    title = {From {Generation} to {Judgment}: {Opportunities} and {Challenges} of {LLM}-as-a-judge},
    shorttitle = {From {Generation} to {Judgment}},
    url = {http://arxiv.org/abs/2411.16594},
    doi = {10.48550/arXiv.2411.16594},
    urldate = {2025-06-11},
    publisher = {arXiv},
    author = {Li, Dawei and Jiang, Bohan and Huang, Liangjie and Beigi, Alimohammad and Zhao, Chengshuai and Tan, Zhen and Bhattacharjee, Amrita and Jiang, Yuxuan and Chen, Canyu and Wu, Tianhao and Shu, Kai and Cheng, Lu and Liu, Huan},
    month = feb,
    year = {2025},
    note = {arXiv:2411.16594 [cs]},
}

@article{spitzer2006brief,
  title={A brief measure for assessing generalized anxiety disorder: the GAD-7},
  author={Spitzer, Robert L and Kroenke, Kurt and Williams, Janet BW and L{\"o}we, Bernd},
  journal={Archives of internal medicine},
  volume={166},
  number={10},
  pages={1092--1097},
  year={2006},
  publisher={American Medical Association}
}

@misc{hopkins_mental_health,
  author = {{Johns Hopkins Medicine}},
  title = {{Mental Health Disorder Statistics}},
  howpublished = {\url{https://www.hopkinsmedicine.org/health/wellness-and-prevention/mental-health-disorder-statistics}},
  year = {2023},
  note = {Accessed: November 14, 2025}
}

@article{sheikh2021wearable,
  title={Wearable, environmental, and smartphone-based passive sensing for mental health monitoring},
  author={Sheikh, Mahsa and Qassem, Meha and Kyriacou, Panicos A},
  journal={Frontiers in digital health},
  volume={3},
  pages={662811},
  year={2021},
  publisher={Frontiers Media SA}
}

@article{jacobson2021deep,
  title={Deep learning paired with wearable passive sensing data predicts deterioration in anxiety disorder symptoms across 17--18 years},
  author={Jacobson, Nicholas C and Lekkas, Damien and Huang, Raphael and Thomas, Natalie},
  journal={Journal of affective disorders},
  volume={282},
  pages={104--111},
  year={2021},
  publisher={Elsevier}
}

@article{nepal2024capturing,
  title={Capturing the college experience: A four-year mobile sensing study of mental health, resilience and behavior of college students during the pandemic},
  author={Nepal, Subigya and Liu, Wenjun and Pillai, Arvind and Wang, Weichen and Vojdanovski, Vlado and Huckins, Jeremy F and Rogers, Courtney and Meyer, Meghan L and Campbell, Andrew T},
  journal={Proceedings of the ACM on interactive, mobile, wearable and ubiquitous technologies},
  volume={8},
  number={1},
  pages={1--37},
  year={2024},
  publisher={ACM New York, NY, USA}
}

@article{gomes2023survey,
  title={A survey on wearable sensors for mental health monitoring},
  author={Gomes, Nuno and Pato, Matilde and Lourenco, Andre Ribeiro and Datia, Nuno},
  journal={Sensors},
  volume={23},
  number={3},
  pages={1330},
  year={2023},
  publisher={MDPI}
}

@article{yoon2024my,
  title={By my eyes: Grounding multimodal large language models with sensor data via visual prompting},
  author={Yoon, Hyungjun and Tolera, Biniyam Aschalew and Gong, Taesik and Lee, Kimin and Lee, Sung-Ju},
  journal={arXiv preprint arXiv:2407.10385},
  year={2024}
}

@article{gruver2023large,
  title={Large language models are zero-shot time series forecasters},
  author={Gruver, Nate and Finzi, Marc and Qiu, Shikai and Wilson, Andrew G},
  journal={Advances in Neural Information Processing Systems},
  volume={36},
  pages={19622--19635},
  year={2023}
}

@article{liu2025feddhad,
  author = {Liu, Ji and Ma, Beichen and Yu, Qiaolin and Jin, Ruoming and Zhou, Jingbo and Zhou, Yang and Dai, Huaiyu and Wang, Haixun and Dou, Dejing and Valduriez, Patrick},
  title = {Efficient Federated Learning with Heterogeneous Data and Adaptive Dropout},
  year = {2025},
  volume = {19},
  number = {8},
  journal = {ACM Trans. Knowl. Discov. Data},
  doi = {10.1145/3749376},
}

@misc{zhou2025april,
  title={APRIL: Active Partial Rollouts in Reinforcement Learning to Tame Long-tail Generation},
  author={Yuzhen Zhou and Jiajun Li and Yusheng Su and Gowtham Ramesh and Zilin Zhu and Xiang Long and Chenyang Zhao and Jin Pan and Xiaodong Yu and Ze Wang and Kangrui Du and Jialian Wu and Ximeng Sun and Jiang Liu and Qiaolin Yu and Hao Chen and Zicheng Liu and Emad Barsoum},
  year={2025},
  eprint={2509.18521},
  archivePrefix={arXiv},
  primaryClass={cs.LG},
  url={https://arxiv.org/abs/2509.18521},
}

@misc{liu2026mosiv,
  title={MOSIV: Multi-Object System Identification from Videos},
  author={Chunjiang Liu and Xiaoyuan Wang and Qingran Lin and Albert Xiao and Haoyu Chen and Shizheng Wen and Hao Zhang and Lu Qi and Ming-Hsuan Yang and Laszlo A. Jeni and Min Xu and Yizhou Zhao},
  year={2026},
  eprint={2603.06022},
  archivePrefix={arXiv},
  primaryClass={cs.CV},
  url={https://arxiv.org/abs/2603.06022},
}

@article{wang2025holigs,
  title={HoliGS: Holistic Gaussian Splatting for Embodied View Synthesis},
  author={Wang, Xiaoyuan and Zhao, Yizhou and Ye, Botao and Shan, Xiaojun and Lyu, Weijie and Qi, Lu and Chan, Kelvin CK and Li, Yinxiao and Yang, Ming-Hsuan},
  journal={arXiv preprint arXiv:2506.19291},
  year={2025}
}

@inproceedings{yao2025measure,
  title={The Measure of All Measures: Quantifying {LLM} Benchmark Quality},
  author={Jihan Yao and Peter Jin and Ke Bao and Qiaolin Yu and Khushi Bhardwaj and Chang Su and Jialei Wang and Yikai Zhu and Sugam Devare and Damon Mosk-Aoyama and Zhen Dong and Venkat Krishna Srinivasan and Yineng Zhang and Oleksii Kuchaiev and Jiantao Jiao and Banghua Zhu},
  booktitle={NeurIPS 2025 Workshop on Evaluating the Evolving LLM Lifecycle},
  year={2025},
  url={https://openreview.net/forum?id=HpnGmTIs7Z}
}

\appendix

\section{Implementation Details}\label{apd:implementation_details}

\textbf{Training Specifications.} Training is conducted on 4×H200 GPUs under a DeepSpeed distributed environment with bf16 precision. We use the AdamW optimizer with a cosine learning-rate schedule. The per-device batch size is set to 4, and gradients are accumulated for 32 steps, resulting in an effective batch size of 128 sequences per update. The model is fine-tuned with a base learning rate of $1\times10^{-5}$.

\textbf{Baseline Inference Configurations.} For all baseline models, inference parameters follow the default configurations provided in the official Qwen2.5 Hugging Face repositories, with temperature set to 0.7, top-$p$ to 0.8, and top-$k$ to 20. Few-shot exemplars are selected from the training split and formatted according to the prompt templates in Appendix~\S\ref{apd:prompt_templates}.

\textbf{Dataset Splits.} All experiments use participant-level splits to prevent information leakage across EMA windows from the same individual. The 258 participants are partitioned into disjoint training (180; 69.77\%), validation (38; 14.73\%), and test (40; 15.50\%) sets, following an approximate 70:15:15 ratio. All windows from a given participant appear in only one split, ensuring evaluation on unseen individuals.

\section{EMA Questions} \label{apd:ema_questions}
The Ecological Momentary Assessment (EMA) used in this study consists of 13 primary items designed to monitor intra-day variations in mental health (Table \ref{tab:ema_question_categories}). These items were specifically adapted from the Patient Health Questionnaire (PHQ)~\citep{kroenke2001phq} and the Generalized Anxiety Disorder (GAD)~\cite{spitzer2006brief} scale to assess symptoms over the preceding four-hour window. As shown in Table 2, the questionnaire covers 14 distinct categories, including core depression indicators like anhedonia and depressed mood, alongside physical and cognitive markers such as somatic discomfort, fatigue, and concentration. It also includes a negative event question to understand context about mental health symptoms.

\begin{table*}[t]
  \centering
  \small 
  \begin{tabularx}{\textwidth}{@{}lX@{}}
    \hline
    \textbf{Question Categories} & \textbf{Questions} \\
    \hline
    1. Anhedonia (Interest/Pleasure) & In the past 4 hours, how much has the user shown little interest or pleasure in activities? \\
    \hline
    2. Depressed Mood & In the past 4 hours, how much has the user appeared down, depressed, or hopeless? \\
    \hline
    3. Sleep Disturbance & Last night, how much trouble did the user have with sleep? \\
    \hline
    4. Fatigue / Energy & In the past 4 hours, how tired or low in energy has the user been? \\
    \hline
    5. Appetite Change & In the past 4 hours, how much has the user shown a poor appetite or overeating? \\
    \hline
    6. Self-worth / Guilt & In the past 4 hours, how much has the user felt bad about themselves? \\
    \hline
    7. Concentration & In the past 4 hours, how much trouble has the user had concentrating? \\
    \hline
    8. Psychomotor Change & In the past 4 hours, how much has the user been moving or speaking more slowly than usual? \\
    \hline
    9. Suicidal Ideation & In the past 4 hours, how often has the user had thoughts of harming themselves or wishing to be dead? \\
    \hline
    10. Somatic Discomfort & In the past 4 hours, how much has the user experienced headache, abdominal discomfort, or body aches? \\
    \hline
    11. Inverted Question & An inverted question randomized from Q1, Q4, or Q7. \\
    \hline
    12. Anxiety Arousal & In the past 4 hours, how much has the user felt nervous, anxious, or on edge? \\
    \hline
    13. Uncontrollable Worry & In the past 4 hours, how much has the user been unable to stop or control worrying? \\
    \hline
    14. Negative Event & In the past 4 hours, did the user experience a negative event? If yes: How negative was the event? \\
    \hline
    Overall Summary & Please summarize the user's overall mental and physical state in the past 4 hours, integrating mood, energy, sleep, appetite, concentration, and physical symptoms. \\
    \hline
  \end{tabularx}
  \caption{\textbf{EMA Questions.} Categories of depression- and anxiety-related symptoms and their corresponding questions or statements.}
  \label{tab:ema_question_categories}
\end{table*}

\section{Sensor Data Preprocessing} \label{apd:sensor_preprocessing}

The sensor streams are preprocessed as follows. For steps and heart rate, we applied empirical thresholds to remove outliers. Stress and accelerometer-derived features were already processed by the data collection software and required no additional filtering. GPS data was reduced to latitude and longitude pairs. Phone lock and unlock state was encoded as a binary sequence, where 0 indicates locked and 1 indicates unlocked, and downsampled from a 1-second to a 60-second resolution to avoid excessively long sequences. Conversation events were logged only when speech was detected, and for each EMA-aligned window we summed event durations to obtain total conversational time in seconds. Sleep duration was recorded once per day as the previous night’s total hours slept.

After preprocessing the sensor streams, we standardized all signals to fixed sampling rates, including heart rate every 10 seconds, GPS every 10 minutes, steps every 60 seconds, stress every 60 seconds, ZCR every 30 seconds, and phone lock and unlock every 60 seconds. Sleep duration and conversation length were stored as scalar values for each window. This unified representation ensured consistent temporal alignment across all sensing modalities.

\section{Metric Definitions} \label{sec:metric_appendix}

\subsection{Coverage}
Presence detection is computed over all symptom categories. Let $a_j,\hat{a}_j\in\{0,1\}$ denote
reference and predicted presence for category $j$. Then:
\begin{align*}
  P_{\mathrm{cov}}&=\frac{TP}{TP+FP},\quad
  R_{\mathrm{cov}}=\frac{TP}{TP+FN},\\
  F1_{\mathrm{cov}}&=\frac{2P_{\mathrm{cov}}R_{\mathrm{cov}}}{P_{\mathrm{cov}}+R_{\mathrm{cov}}},
\end{align*}
where $TP$, $FP$, and $FN$ represent correct mentions, hallucinations, and omissions.

\subsection{Presence-aware Severity Alignment}
Severity is scored only when either reference or prediction marks a symptom as present:
\[
  \mathcal{J}=\{j \mid a_j=1\ \text{or}\ \hat a_j=1\}.
\]
For each $j\in\mathcal{J}$, a weight $w_j$ is assigned based on ordinal deviation:
\[
  w_j=
  \begin{cases}
    0, & a_j\neq \hat a_j, \\
    1, & |s_j-\hat{s}_j|=0, \\
    0.75, & |s_j-\hat{s}_j|=1, \\
    0.25, & |s_j-\hat{s}_j|=2, \\
    0, & |s_j-\hat{s}_j|\ge 3,
  \end{cases}
\]
and the final alignment score is
\[
  S_{\mathrm{align}}=\frac{1}{|\mathcal{J}|}\sum_{j\in\mathcal{J}} w_j.
\]
This captures both hallucination penalties ($a_j\neq \hat a_j$) and ordinal severity mismatch.

\section{Additional Results} \label{apd:additional_results}

\begin{table*}[t]
  \centering
  \scriptsize
  \setlength{\tabcolsep}{5pt} 
  \renewcommand{\arraystretch}{1.25}

  \begin{adjustbox}{max width=\textwidth}
    \begin{tabular}{
        l l
        c c c c c c
        c c
      }
      \toprule
      \multicolumn{2}{c}{\textbf{Model Specification}} &
      \multicolumn{6}{c}{\textbf{Linguistic Metrics}} &
      \multicolumn{2}{c}{\textbf{LLM-as-a-Judge Metrics}}\\
      \cmidrule(lr){1-2}
      \cmidrule(lr){3-8}
      \cmidrule(lr){9-10}

      \textbf{Method} &
      \textbf{Model} &
      \textbf{ROUGE-1} &
      \textbf{ROUGE-2} &
      \textbf{ROUGE-L} &
      \textbf{BLEU-4} &
      \textbf{METEOR} &
      \textbf{BERTScore} &
      \textbf{Coverage} &
      \textbf{Alignment}\\ 
      \midrule


      \multicolumn{10}{c}{\textit{Summary-level Evaluation}}\\[2pt]
      \midrule
      LENS &
      Qwen2.5-14B-Instruct &
      \textbf{0.593} & \textbf{0.265} & \textbf{0.410} &
      \textbf{0.220} & \textbf{0.467} & \textbf{0.776} &
      0.753 & 0.601 \\

      TS-Text &
      Qwen2.5-14B-Instruct &
      0.294 & 0.057 & 0.151 &
      0.032 & 0.218 & 0.631 &
      0.614 & 0.372 \\

      TS-Image &
      Qwen2.5-VL-32B &
      0.556 & 0.212 & 0.373 &
      0.166 & 0.452 & 0.765 &
      0.740 & 0.579 \\
      \midrule


      \multicolumn{10}{c}{\textit{Item-level Evaluation}}\\[2pt]
      \midrule
      LENS &
      Qwen2.5-14B-Instruct &
      0.624 & 0.460 & 0.603 &
      0.383 & 0.611 & 0.832 &
      - & 0.732 \\ 

      TS-Text &
      Qwen2.5-14B-Instruct &
      0.176 & 0.047 & 0.142 &
      0.017 & 0.218 & 0.604 &
      - & 0.665 \\

      TS-Image &
      Qwen2.5-VL-32B &
      0.162 & 0.044 & 0.136 &
      0.016 & 0.211 & 0.602 &
      - & 0.687 \\
      \bottomrule
    \end{tabular}
  \end{adjustbox}
  \caption{\textbf{Evaluation on summary-level and item-level QA datasets across all linguistic and clinical alignment metrics.} LENS demonstrates consistent superiority in both narrative generation and specific symptom retrieval.}
  \label{tab:narrative_comparison}
\end{table*}

\begin{table*}[]
  \centering
  \scriptsize
  \setlength{\tabcolsep}{5pt}
  \renewcommand{\arraystretch}{1.25}

  \begin{adjustbox}{max width=\textwidth}
    \begin{tabular}{
        l l
        c c c c c c
        c c
      }
      \toprule
      \multicolumn{2}{c}{\textbf{Model Specification}} &
      \multicolumn{6}{c}{\textbf{Linguistic Metrics}} &
      \multicolumn{2}{c}{\textbf{LLM-as-a-Judge Metrics}}\\
      \cmidrule(lr){1-2}
      \cmidrule(lr){3-8}
      \cmidrule(lr){9-10}

      \textbf{Method} &
      \textbf{Modality} &
      \textbf{ROUGE-1} &
      \textbf{ROUGE-2} &
      \textbf{ROUGE-L} &
      \textbf{BLEU-4} &
      \textbf{METEOR} &
      \textbf{BERTScore} &
      \textbf{Coverage} &
      \textbf{Alignment}\\
      \midrule


      \multicolumn{10}{c}{\textit{Summary-level Evaluation}}\\[2pt]
      \midrule

      LENS &
      Multimodal &
      \textbf{0.593} & 0.265 & \textbf{0.410} &
      \textbf{0.220} & \textbf{0.467} & \textbf{0.776} &
      0.753 & \textbf{0.601} \\

      TS-Text-FT  &
      Text-Only &
      0.578 & \textbf{0.275} & 0.390 &
      0.207 & 0.408 & 0.772 &
      \textbf{0.789} & 0.583 \\

      \midrule


      \multicolumn{10}{c}{\textit{Item-level Evaluation}}\\[2pt]
      \midrule

      LENS &
      Multimodal &
      \textbf{0.624} & \textbf{0.460} & \textbf{0.603} &
      \textbf{0.383} & \textbf{0.611} & \textbf{0.832} &
      \multicolumn{1}{c}{-} & \textbf{0.732} \\

      TS-Text-FT  &
      Text-Only &
      0.192 & 0.078 & 0.172 &
      0.032 & 0.281 & 0.598 &
      \multicolumn{1}{c}{-} & - \\

      \bottomrule
    \end{tabular}
  \end{adjustbox}
  \caption{\textbf{Ablation Study: with or without time series encoder.}  Comparison between LENS and TS-Text-FT  (fine-tuned on text-serialized time series).}
  \label{tab:ablation_textft}
\end{table*}

\noindent \textbf{Impact of Time Series Modality.}
To disentangle the effect of supervised fine-tuning from the contribution of native time-series modeling, we conduct an ablation using a text-only baseline (\textbf{TS-Text-FT}). In this setting, we remove the time-series encoder from LENS and fine-tune the same Qwen2.5-14B backbone on the identical sensor–text training pairs. Instead of being encoded by a patch-based time-series encoder, sensor streams are directly serialized as textual inputs. To control for sequence truncation effects, the text-only model is trained with a cutoff length of 20{,}000 tokens, covering more than 99.95\% of training samples.

As reported in Table~\ref{tab:ablation_textft}, LENS consistently outperforms TS-Text-FT across all evaluation dimensions. For summary-level generation, LENS achieves higher clinical alignment (0.601 vs.\ 0.583) and stronger linguistic quality (METEOR score of 0.467 vs.\ 0.408). The gap widens substantially in the item-level QA setting, where TS-Text-FT exhibits difficulty reasoning over long, unstructured numerical sequences, resulting in markedly lower ROUGE-L scores (0.172 vs.\ 0.603). These results indicate that supervised fine-tuning alone is insufficient to capture the temporal structure of sensor data, and that explicit time-series encoding plays a central role in LENS’s performance.

\paragraph{Computational Efficiency and Token Consumption.}
In addition to accuracy, we analyze the computational efficiency of different modeling paradigms by comparing their token consumption. Specifically, we measure the mean prefill token count per sample and the total token consumption over the Narrative ($n=8{,}192$) and Item-level QA ($n=16{,}384$) datasets for three approaches: text-only (Qwen2.5), vision-based (Qwen2.5-VL-32B), and LENS.

For LENS, effective prefill tokens are computed as the sum of textual tokens and patch-based embeddings (patch size $k=8$) that replace raw time-series placeholders. For the vision-language baseline, token counts follow the model’s native processor, which accounts for both textual inputs and image-derived tokens. For the text-only baseline, the tokenizer needs to handle all the sensor streams as text, resulting in substantially longer input sequences.

\begin{figure}[t]
  \centering
  \includegraphics[width=1\linewidth]{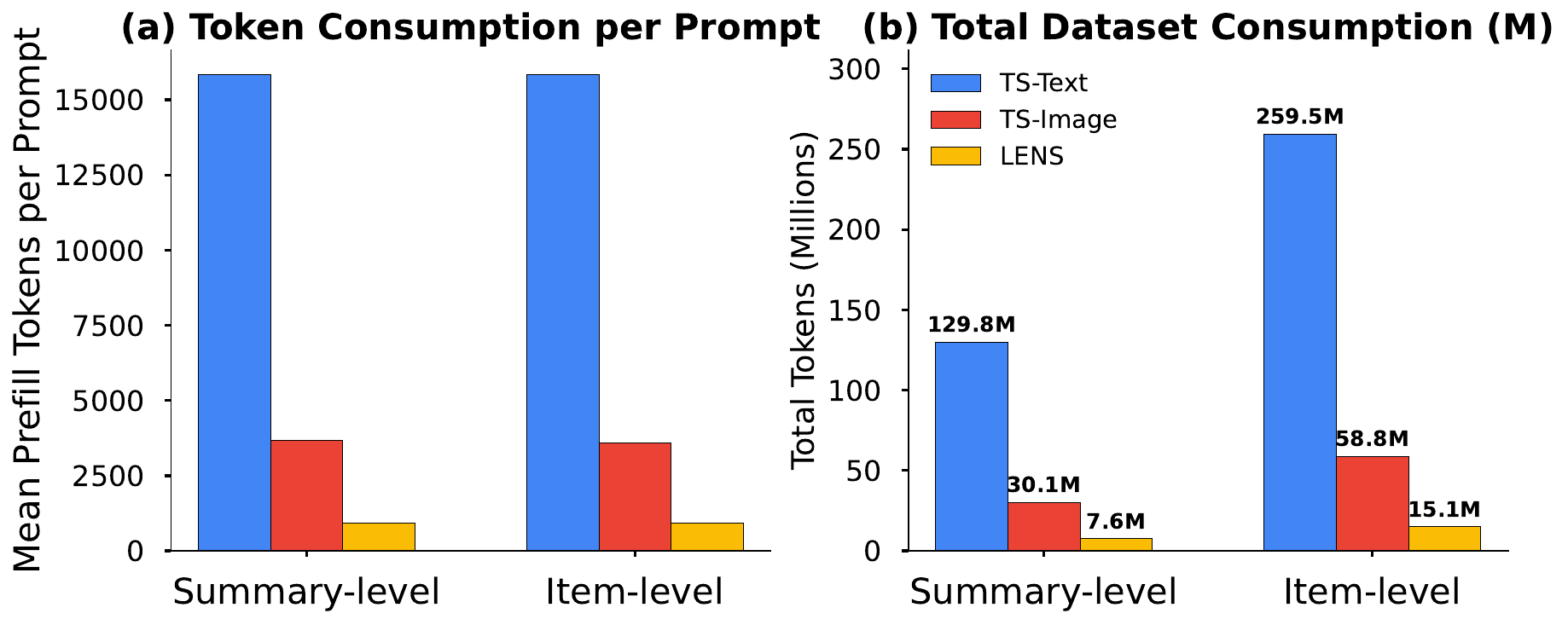}
  \caption{\textbf{Computational Efficiency Analysis.} Comparison of token consumption across three modalities: Text (Qwen-2.5), Vision-Language (TS-Image/Qwen2.5-VL), and Time-Series (LENS). \textbf{(a)} Mean prefill tokens per prompt. \textbf{(b)} Total dataset token consumption (in Millions) for Narrative and QA datasets.}
  \label{fig:token_consumption}
\end{figure}

Figure~\ref{fig:token_consumption} summarizes the results. Across both tasks, LENS exhibits the lowest token consumption by a large margin. Compared to the text-only baseline, which averages over 15{,}800 tokens per sample due to verbose serialization of high-frequency signals, LENS requires approximately 930 tokens per sample, corresponding to a reduction of about 94\%. Relative to the vision-language model, which represents signals as plots, LENS remains roughly four times more token-efficient (933 vs.\ 3{,}679 tokens).

This reduction in context length provides a practical explanation for the degraded QA performance observed in text-only fine-tuning, where long serialized inputs are prone to attention dilution in extended contexts. By encoding raw signals into compact patch-level representations, LENS substantially reduces sequence length while preserving task-relevant temporal information, enabling more efficient and scalable inference for long-duration health monitoring scenarios.

\noindent \textbf{Impact of Data Scale.}
To investigate the influence of training data volume, we trained LENS using varying proportions (10\%, 50\%, and 100\%) of the generated dataset, keeping the base model architecture consistent. As shown in Table \ref{tab:data_efficiency}, the impact of data scale varies across different evaluation dimensions. For summary-level narrative generation, Presence Alignment improves strictly monotonically with data size, rising from 0.545 (10\% data) to 0.601 (100\% data), indicating that larger datasets are essential for minimizing hallucinations in complex clinical summaries. However, Symptom Coverage is not strictly monotonic; the model trained on 50\% data achieved the highest coverage (0.823), slightly outperforming the full model (0.801) and the 10\% model (0.783). In the Item-level QA task, the full 100\% model achieves the best overall performance in both linguistic metrics (ROUGE-L 0.603) and clinical alignment (0.732). Notably, the model trained on only 10\% data exhibits surprisingly strong performance in QA tasks, achieving a higher alignment score (0.729) than the 50\% model (0.706). This suggests that while basic instruction-following for short-form QA can be established with smaller data scales, the full dataset is necessary to maximize structural coherence and clinical fidelity in longer narratives.

\noindent\textbf{Smaller LENS Variants Remain Competitive.} To address computational feasibility for deployment in resource-constrained environments, we evaluate LENS-7B—a lighter variant of our framework—alongside models trained on reduced data proportions. As shown in our comparison, LENS-7B achieves slightly lower performance on complex clinical alignment than LENS-14B (0.559 vs. 0.601), reflecting the trade-off between model size and reasoning depth. Nevertheless, LENS-7B matches the 14B model's mean performance on linguistic metrics (e.g., ROUGE-L 0.409 vs. 0.410) and closely trails it in Item-level QA tasks (Presence Alignment 0.727 vs. 0.732). Similarly, models trained on limited data demonstrate surprising robustness in short-form generation. These results suggest that LENS-7B or low-data fine-tuning provides a strong balance between efficiency and performance, making it a cost-effective choice for real-world, resource-limited scenarios such as edge-device deployment.

\begin{table*}[]
  \centering
  \scriptsize
  \setlength{\tabcolsep}{5pt}
  \renewcommand{\arraystretch}{1.25}

  \begin{adjustbox}{max width=\textwidth}
    \begin{tabular}{
        l l
        c c c c c c
        c c
      }
      \toprule
      \multicolumn{2}{c}{\textbf{Model Specification}} &
      \multicolumn{6}{c}{\textbf{Linguistic Metrics}} &
      \multicolumn{2}{c}{\textbf{LLM-as-a-Judge Metrics}}\\
      \cmidrule(lr){1-2}
      \cmidrule(lr){3-8}
      \cmidrule(lr){9-10}

      \textbf{Method} &
      \textbf{Data Size} &
      \textbf{ROUGE-1} &
      \textbf{ROUGE-2} &
      \textbf{ROUGE-L} &
      \textbf{BLEU-4} &
      \textbf{METEOR} &
      \textbf{BERTScore} &
      \textbf{Coverage} &
      \textbf{Alignment}\\
      \midrule


      \multicolumn{10}{c}{\textit{Summary-level Evaluation}}\\[-2pt]
      \midrule

      LENS &
      100\% Data &
      \textbf{0.593} & \textbf{0.265} & \textbf{0.410} &
      \textbf{0.220} & 0.467 & \textbf{0.776} &
      0.801 & \textbf{0.601} \\

      LENS &
      50\% Data &
      0.586 & 0.255 & 0.399 &
      0.211 & 0.463 & 0.773 &
      \textbf{0.823} & 0.585 \\

      LENS &
      10\% Data &
      0.583 & 0.260 & 0.400 &
      0.213 & \textbf{0.475} & 0.772 &
      0.783 & 0.545 \\
      \midrule


      \multicolumn{10}{c}{\textit{Item-level Evaluation}}\\[-2pt]
      \midrule

      LENS &
      100\% Data &
      \textbf{0.624} & \textbf{0.460} & \textbf{0.603} &
      \textbf{0.383} & \textbf{0.611} & \textbf{0.832} &
      \multicolumn{1}{c}{-} & \textbf{0.732} \\

      LENS &
      50\% Data &
      0.611 & 0.446 & 0.590 &
      0.369 & 0.603 & 0.823 &
      \multicolumn{1}{c}{-} & 0.706 \\

      LENS &
      10\% Data &
      0.612 & 0.451 & 0.594 &
      0.371 & 0.602 & 0.827 &
      \multicolumn{1}{c}{-} & 0.729 \\

      \bottomrule
    \end{tabular}
  \end{adjustbox}
  \caption{\textbf{Data Efficiency Analysis.} Performance comparison of LENS trained on varying proportions of the dataset (100\% vs. 50\% vs. 10\%). Results are reported for both Summary-level (narrative generation) and Item-level (QA) tasks.}
  \label{tab:data_efficiency}
\end{table*}
\begin{table*}[h!]
  \centering
  \scriptsize
  \setlength{\tabcolsep}{5pt}
  \renewcommand{\arraystretch}{1.25}

  \begin{adjustbox}{max width=\textwidth}
    \begin{tabular}{
        l l
        c c c c c c
        c c
      }
      \toprule
      \multicolumn{2}{c}{\textbf{Model Specification}} &
      \multicolumn{6}{c}{\textbf{Linguistic Metrics}} &
      \multicolumn{2}{c}{\textbf{LLM-as-a-Judge Metrics}}\\
      \cmidrule(lr){1-2}
      \cmidrule(lr){3-8}
      \cmidrule(lr){9-10}

      \textbf{Method} &
      \textbf{Model Size} &
      \textbf{ROUGE-1} &
      \textbf{ROUGE-2} &
      \textbf{ROUGE-L} &
      \textbf{BLEU-4} &
      \textbf{METEOR} &
      \textbf{BERTScore} &
      \textbf{Coverage} &
      \textbf{Alignment}\\
      \midrule


      \multicolumn{10}{l}{\textit{Summary-level Evaluation}}\\[2pt]
      \midrule

      LENS &
      14B &
      \textbf{0.593} & \textbf{0.265} & \textbf{0.410} &
      \textbf{0.220} & 0.467 & \textbf{0.776} &
      0.753 & \textbf{0.601} \\

      LENS &
      7B &
      0.592 & 0.265 & 0.409 &
      0.220 & \textbf{0.468} & 0.775 &
      \textbf{0.826} & 0.559 \\

      \midrule


      \multicolumn{10}{l}{\textbf{Item-level Evaluation}}\\[2pt]
      \midrule

      LENS &
      14B &
      \textbf{0.624} & \textbf{0.460} & \textbf{0.603} &
      \textbf{0.383} & \textbf{0.611} & \textbf{0.832} &
      \multicolumn{1}{c}{-} & \textbf{0.732} \\

      LENS &
      7B &
      0.622 & 0.457 & 0.600 &
      0.380 & 0.608 & 0.831 &
      \multicolumn{1}{c}{-} & 0.727 \\

      \bottomrule
    \end{tabular}
  \end{adjustbox}
  \caption{\textbf{Model Size Analysis.} Performance comparison between LENS-14B and LENS-7B across summary-level and item-level tasks.}
  \label{tab:model_size_comparison}
\end{table*}

\noindent\textbf{Encoder Hyperparameter Sensitivity.}
We ablate the MLP depth ($L\in\{3,5,7\}$) and patch size ($p\in\{4,8,16\}$) around the default ($L{=}5$, $p{=}8$) using the 10\% data setting (Table~\ref{tab:data_efficiency}). As shown in Table~\ref{tab:ablation_encoder}, all metrics fluctuate within 0.01 absolute, confirming that LENS is robust to encoder hyperparameter choices.

\begin{table}[h]
  \centering
  \resizebox{\columnwidth}{!}{
    \begin{tabular}{@{}llcccc@{}}
      \toprule
      \textbf{Ablation} & \textbf{Setting} & \textbf{R-L} & \textbf{BERTSc.} & \textbf{Cov.} & \textbf{Align.} \\
      \midrule
      \multicolumn{6}{c}{\textit{Summary-level}} \\
      \midrule
      \multirow{3}{*}{Depth ($L$)}
      & $L{=}3$              & 0.400 & 0.771 & 0.780 & 0.553 \\
      & $L{=}5$ \textbf{(D)} & 0.400 & 0.772 & 0.783 & 0.545 \\
      & $L{=}7$              & 0.399 & 0.772 & 0.789 & 0.565 \\
      \cmidrule(l){1-6}
      \multirow{3}{*}{Patch ($p$)}
      & $p{=}4$              & 0.406 & 0.775 & 0.797 & 0.574 \\
      & $p{=}8$ \textbf{(D)} & 0.400 & 0.772 & 0.783 & 0.545 \\
      & $p{=}16$             & 0.401 & 0.773 & 0.787 & 0.562 \\
      \midrule
      \multicolumn{6}{c}{\textit{Item-level}} \\
      \midrule
      \multirow{3}{*}{Depth ($L$)}
      & $L{=}3$              & 0.599 & 0.829 & -- & 0.740 \\
      & $L{=}5$ \textbf{(D)} & 0.594 & 0.827 & -- & 0.729 \\
      & $L{=}7$              & 0.595 & 0.828 & -- & 0.733 \\
      \cmidrule(l){1-6}
      \multirow{3}{*}{Patch ($p$)}
      & $p{=}4$              & 0.607 & 0.834 & -- & 0.734 \\
      & $p{=}8$ \textbf{(D)} & 0.594 & 0.827 & -- & 0.729 \\
      & $p{=}16$             & 0.603 & 0.830 & -- & 0.735 \\
      \bottomrule
    \end{tabular}
  }
  \caption{\textbf{Encoder Hyperparameter Ablation.} \textbf{(D)} marks the default. Evaluated under the 10\% data setting. R-L = ROUGE-L; BERTSc. = BERTScore; Cov. = Coverage; Align. = Presence Alignment.}
  \label{tab:ablation_encoder}
\end{table}

\noindent\textbf{Sensor Mismatch Experiment.}
To empirically test whether LENS relies on actual sensor signals for clinical content generation, we conducted a sensor mismatch experiment on the full narrative test set. For each sample, sensor streams were randomly shuffled across participants while questions and ground-truth labels remained fixed. If LENS had learned only stylistic patterns from the LLM-generated training labels, performance should remain stable under shuffling; if it actively grounds clinical content in sensor signals, Presence Alignment should degrade.

As shown in Table~\ref{tab:sensor_mismatch}, linguistic metrics (ROUGE-L, BERTScore) remain nearly unchanged under sensor shuffling, consistent with partial style learning from the fine-tuning process. In contrast, Presence Alignment decreases by 12.1 percentage points (a 20.1\% relative decline), indicating that correct identification of symptom presence and ordinal severity depends on the actual sensor signals at inference time. This dissociation provides direct evidence that LENS does not generate fixed template-based outputs independent of sensor input, but actively uses sensor signals to ground clinical content.

\begin{table}[h]
  \centering
  \resizebox{\columnwidth}{!}{
    \begin{tabular}{@{}lccc@{}}
      \toprule
      \textbf{Condition} & \textbf{ROUGE-L} & \textbf{BERTScore} & \textbf{Presence Align.} \\
      \midrule
      Original        & 0.410 & 0.776 & 0.601 \\
      Sensor Mismatch & 0.408 & 0.775 & 0.480 \\
      \midrule
      $\Delta$        & $-$0.002 & $-$0.001 & $-$0.121 (20.1\%) \\
      \bottomrule
    \end{tabular}
  }
  \caption{\textbf{Sensor Mismatch Experiment.} Shuffling sensor streams across participants leaves linguistic metrics nearly unchanged but causes a 20.1\% relative decline in Presence Alignment, confirming that LENS actively grounds clinical content in sensor signals.}
  \label{tab:sensor_mismatch}
\end{table}

\noindent\textbf{Dataset Construction Cost.}
Table~\ref{tab:cost_analysis} reports the computational cost of the LENS dataset construction pipeline, estimated using commercial API pricing as a conservative upper bound. All models were run locally on institutional GPU clusters via SGLang, incurring zero actual API charges.

\begin{table}[h]
  \centering
  \resizebox{\columnwidth}{!}{
    \begin{tabular}{@{}llrrc@{}}
      \toprule
      \textbf{Stage} & \textbf{Model} & \textbf{In Tok.} & \textbf{Out Tok.} & \textbf{Cost} \\
      \midrule
      Narrative Enrich. & Qwen2.5-14B  & 22.71M & 10.34M & \$3.30 \\
      QA Enrichment     & Qwen2.5-14B  & 30.15M &  1.76M & \$3.19 \\
      LLM-as-Judge $\times$3 & 7--8B models & 16.83M &  3.00M & \$1.25 \\
      \midrule
      \textbf{Total}    &              & \textbf{69.7M} & \textbf{15.1M} & \textbf{\$7.75} \\
      \bottomrule
    \end{tabular}
  }
  \caption{\textbf{Dataset Construction Cost Analysis.} Estimated cost using commercial API pricing. The pipeline is used only for offline dataset construction; inference requires only the encoder and LLM backbone.}
  \label{tab:cost_analysis}
\end{table}

\noindent\textbf{PHQ-9 Symptom Distribution.}
Table~\ref{tab:phq9_distribution} reports the ordinal response distributions across all nine PHQ-9 items. Despite recruiting exclusively from a clinical MDD cohort, responses span the full 0--3 severity range with substantial representation at every level, reflecting natural within-cohort symptom variability rather than annotation bias.

\begin{table}[h]
  \centering
  \resizebox{\columnwidth}{!}{
    \begin{tabular}{@{}lcccc@{}}
      \toprule
      \textbf{Item} & \textbf{Not at all} & \textbf{Sometimes} & \textbf{Often} & \textbf{Constantly} \\
      \midrule
      Q1 Anhedonia       & 21.5\% & 27.3\% & 30.2\% & 21.1\% \\
      Q2 Depressed Mood  & 23.8\% & 26.3\% & 29.6\% & 20.3\% \\
      Q3 Sleep Disturb.  & 27.8\% & 21.4\% & 27.2\% & 23.5\% \\
      Q4 Fatigue         & 11.2\% & 17.4\% & 35.5\% & 35.9\% \\
      Q5 Appetite        & 35.2\% & 23.1\% & 24.8\% & 16.9\% \\
      Q6 Self-worth      & 26.0\% & 22.5\% & 30.8\% & 20.8\% \\
      Q7 Concentration   & 22.2\% & 22.5\% & 33.6\% & 21.8\% \\
      Q8 Psychomotor     & 56.3\% & 20.6\% & 15.4\% &  7.8\% \\
      Q9 Suicidal Id.    & 91.3\% &  4.3\% &  2.0\% &  2.5\% \\
      \bottomrule
    \end{tabular}
  }
  \caption{\textbf{PHQ-9 Ordinal Response Distribution.} Raw EMA ground-truth responses ($N{=}50{,}957$ per item). The ``Not at all'' column ranges from 11.2\% to 91.3\%, demonstrating substantial within-cohort variability across all severity levels.}
  \label{tab:phq9_distribution}
\end{table}

\section{User Study}

\noindent\textbf{Survey Design.} The survey was designed with input from both a clinical psychologist and a therapist. Each rating example includes a narrative presented alongside a ground-truth symptom table, followed by questions assessing accuracy and comprehensiveness. Note that the user study excludes the inverted question (Q11 in Table \ref{tab:ema_question_categories}), which is a semantic reversal of an existing item and Sleep Disturbance question which is administered only in morning EMAs to ensure non-redundant and temporally consistent evaluation. Because our expert reviewers noted that mentally tracking symptoms across the narrative was difficult, we incorporated two design changes: (1) presenting the narrative and symptom table simultaneously on the screen, and (2) adding Yes/No checkboxes to help raters track individual symptoms when answering these questions. For the clinical utility and language cohesion items, we display the narrative again to remind raters of the content. After receiving positive design feedback from our expert reviewers, we administered it to the 13 mental health experts. The participants are recruited through the UpWork platform and community therapist forums, they are payed their hourly rates ranging between \$50 to \$150.

\noindent \textbf{Pre-survey Questions.} Before beginning the rating process, raters were asked to answer four background questions:

\begin{enumerate}
  \item What is your professional role or background? (Options: Psychiatrist, Clinical Psychologist, Therapist, Other)
  \item What is your highest degree held? (Options: High school/Diploma, Bachelor’s, Master’s, Doctorate/MD)
  \item How familiar are you with the Patient Health Questionnaire-9 (PHQ-9) for depression screening?
  \item How familiar are you with the Generalized Anxiety Disorder Questionnaire (GAD-7) for anxiety screening?
\end{enumerate}

For Questions 3 and 4, the response options were: Not familiar at all, Slightly familiar, Moderately familiar, and Very familiar. The experts’ pre-survey responses are summarized in Table~\ref{tab:expert_background}.

\begin{table}[]
  \resizebox{\columnwidth}{!}{%
    \begin{tabular}{@{}lllll@{}}
      \toprule
      \textbf{\#} & \textbf{Background} & \textbf{Education} & \textbf{PHQ Familiarity} & \textbf{GAD Familiarity} \\
      \midrule
      1  & Therapist            & Master's               & Very Familiar                   & Very Familiar                   \\
      2 &Psychologist &Doctorate &Very Familiar &Very Familiar \\
      3 & Therapist & Bachelor's &Slightly Familiar & Moderately Familiar \\
      4 & Therapist &Master's &Very Familiar & Very Familiar \\
      5 & Health Coach & Doctorate &Slightly Familiar &Moderately Familiar \\
      6 & Psychologist &Bachelor's &Moderately Familiar &Moderately Familiar \\
      7 &Researcher &Doctorate &Very Familiar & Moderately Familiar \\
      8 &Psychologist &Bachelor's &Moderately Familiar & Moderately Familiar \\
      9 &Psychologist & Master's &Very Familiar &Very Familiar \\
      10 &Psychologist &Master's &Slightly Familiar &Slightly Familiar \\
      11 &Therapist &Master's & Very Familiar & Very Familiar \\
      12 &Therapist & Master's & Moderately Familiar & Very Familiar \\
      13 &Psychologist &Bachelor's &Moderately Familiar &Moderately Familiar \\
      \bottomrule
    \end{tabular}
  }
  \caption{\textbf{Background information of mental health experts.}}
  \label{tab:expert_background}
\end{table}
\label{apd:user_study}

\noindent\textbf{Survey Flow.} After completing the pre-survey questions, each expert rater is introduced to the task through a detailed explanation of the overall goal, problem setting, and rating procedure. They are then shown an annotated example that includes sample answers and reasoning to illustrate how the evaluation should be performed. In every survey, the first example presented is a dummy item intended solely to help raters become familiar with the interface and question format. This is followed by the nine narratives that constitute the actual rating set.

\section{Prompt Templates and Selected Examples}
\label{apd:prompt_templates}

\paragraph{Prompt Templates.}
Tables~\ref{tab:template_for_narrative_rewriting}-\ref{tab:text-qa-prompt} summarize the prompt templates used throughout our experiments for narrative generation, question answering, and evaluation. Tables~\ref{tab:template_for_narrative_rewriting} and~\ref{tab:template_evaluation_quality} define the prompts for rewriting rule-based EMA-derived symptom descriptions into fluent narrative labels and for LLM-based quality assessment of generated narratives. Tables~\ref{tab:structured-eval-prompt} and~\ref{tab:qa-eval-prompt} specify structured evaluation prompts for extracting symptom presence and severity, as well as for assessing ordinal severity alignment in single-item QA. Tables~\ref{tab:vlm-narrative-prompt} and~\ref{tab:vlm-qa-prompt} present the prompts used for vision--language baselines, where multivariate sensor streams are provided as multi-panel time-series visualizations together with contextual features. Finally, Tables~\ref{tab:text-narrative-prompt} and~\ref{tab:text-qa-prompt} describe the text-based baseline prompts, in which the same sensor data are serialized as numerical sequences using placeholder tokens.
We adopt few-shot prompting for non-fine-tuned baselines to mitigate systematic mismatches in output format observed under zero-shot settings; as shown in Table~\ref{tab:zero_shot_example}, zero-shot baselines often fail to follow the required PHQ-9--style narrative structure and instead produce generic, data-centric descriptions. By applying few-shot exemplars to the baseline models, we raise their performance floor and isolate differences arising from input representation and architectural design rather than prompt or format misalignment.

\begin{figure*}[t]
    \centering
    \includegraphics[
      width=\textwidth,
      height=0.85\textheight,
      keepaspectratio
    ]{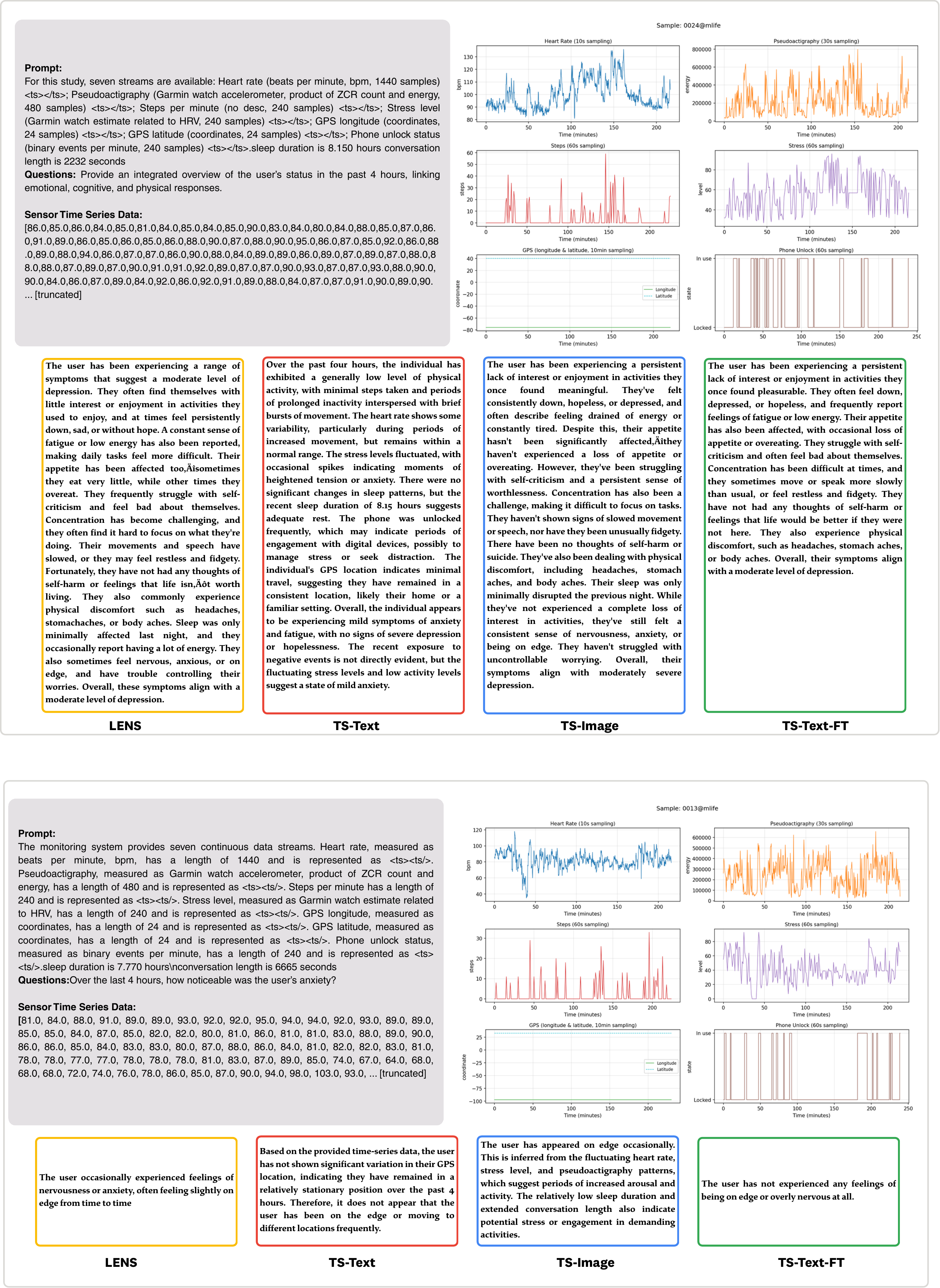}
    \caption{\textbf{Qualitative examples for narrative QA (top) and single-question QA (bottom).} Each example shows the full prompt context, the form of sensor input provided to the model, and the generated output. LENS directly consumes raw multivariate time-series via a patch-based time-series encoder. VLM-based baselines receive the same signals rendered as multi-panel plots, while text-based baselines process serialized numerical sequences. All models are prompted with identical task instructions, enabling a controlled comparison of qualitative behavior across input representations.}
    \label{fig:qualitative_examples}
  \end{figure*}

\paragraph{Selected Qualitative Examples.}
Using the prompt templates described above, we present selected qualitative examples illustrating narrative QA and single-question QA across different model variants (See Figure\ref{fig:qualitative_examples}. Each example shows the prompt context, the form of sensor input provided to the model, and the resulting output. For VLM-based baselines, time-series inputs are rendered as visual plots, whereas LENS directly processes the same raw sensor streams via a patch-based time-series encoder. Text-based baselines receive identical information serialized as numerical arrays.

  \begin{table}[htbp!]
    \centering
    \small 
    \begin{tabularx}{\columnwidth}{|X|}
      \hline
      \rowcolor{gray!15}\textbf{System Prompt} \\
      \hline
      You are both a mental health and language specialist experienced with clinical records concerning mental health conditions.
      Rewrite rule-based psychological assessment templates into fluent, engaging narrative passages, strictly preserving every factual detail and original severity description. These improved narratives serve as ground-truth labels for training AI models to predict mental health states using physiological sensor data (such as smartwatch readings).

      \textbf{Paraphrasing Guidelines}

      1. \textit{Factual Accuracy and Preservation.} Retain every original severity level exactly as presented. Do not add any interpretations, clinical reasoning, or extra context. Preserve all frequency and intensity information without omission or alteration.

      2. \textit{Natural, Readable Language.} Remove mechanical or repetitive phrasing. Employ varied sentence structures and natural transitions between symptoms. Ensure that the narrative reads as a natural human description.

      3. \textit{Consistency in Terminology and Tone.} Use identical language for identical severity levels across all narratives. Maintain a uniform style and tone throughout all paraphrased outputs.

      4. \textit{Accessibility and Clarity.} Write in straightforward, accessible language suitable for general audiences. Avoid technical or clinical terms whenever possible. Use person-first, stigma-free wording. Keep sentences clear, concise, and complete.
      \\
      \hline
      \rowcolor{gray!15}\textbf{User Prompt Template} \\
      \hline
      Your task: Transform the below rule-based assessment into a well-structured, fluent narrative that fully preserves all factual content and improves readability.

      Original Assessment: \{rule\_based\_template\}

      Enhanced Narrative:
      \\
      \hline
    \end{tabularx}
    \caption{\textbf{Prompt template for narrative rewriting.} This prompt is used to rewrite rule-based EMA-derived symptom descriptions into fluent narrative labels.}
    \label{tab:template_for_narrative_rewriting}
  \end{table}

  \onecolumn
  \newpage
  \small
  \begin{longtable}{|>{\raggedright\arraybackslash}p{0.95\columnwidth}|}
    \hline
    \rowcolor{gray!15}\textbf{System Prompt} \\
    \hline
    As a highly meticulous and objective clinical quality reviewer, your primary responsibility is to evaluate the quality and safety of an AI-generated mental health narrative.
    You must ground your judgment strictly in the provided source data.
    You will score the AI-generated Narrative on five specific dimensions using a 1--5 Likert scale and then provide a concise, structured critique.

    \textbf{Template-Based Narrative (Baseline for Comparison/ground):}
    This is the rule-based description generated directly from the PHQ-9 scores using a template.
    Use this as a baseline for factual alignment and coverage assessment.

    \textbf{AI-generated Narrative (To Be Evaluated):}
    This is the AI-rewritten version of the narrative that you must score and critique.
    You must respond with only a valid JSON object, with no additional text before or after.
    \\
    \hline
    \rowcolor{gray!15}\textbf{User Prompt Template} \\
    \hline
    \textbf{Template-Based Narrative (Baseline for Comparison/ground):}

    \{original\_template\}

    \textbf{AI-generated Narrative (To Be Evaluated):}

    \{enriched\_narrative\}

    Please evaluate the AI-generated Narrative based on the following five dimensions.
    For each dimension, provide a score from 1 (Very Poor) to 5 (Excellent).

    \textbf{Factual Alignment:}

    Does the narrative accurately reflect the presence or absence of symptoms reported in the Template-Based Narrative?
    Does it contradict any facts from the source data?

    \textit{Scoring guide:}
    1 indicates significant factual contradictions.
    3 indicates general alignment with minor inaccuracies.
    5 indicates perfect alignment with no factual errors.

    \textbf{Symptom Coverage:}

    Does the narrative mention or allude to all relevant symptoms that were reported with a non-zero score?

    \textit{Scoring guide:}
    1 indicates multiple significant symptoms are missed.
    3 indicates most severe symptoms are covered with some omissions.
    5 indicates comprehensive coverage of all reported symptoms.

    \textbf{Severity Fidelity:}

    Does the language and tone accurately reflect the severity levels from the Template-Based Narrative
    (for example, not at all, several days, more than half the days, nearly every day)?

    \textit{Scoring guide:}
    1 indicates gross misrepresentation of severity.
    3 indicates approximate severity with limited precision.
    5 indicates precise and appropriate severity representation.

    \textbf{Fluency and Naturalness:}

    Is the narrative coherent, well-written, and natural-sounding?
    Does it avoid robotic or repetitive phrasing without sounding artificial?

    \textit{Scoring guide:}
    1 indicates awkward or highly artificial text.
    3 indicates generally fluent but slightly unnatural phrasing.
    5 indicates natural, engaging, and human-like language.

    \textbf{Hallucination Risk:}

    Does the narrative introduce any new symptoms, details, or assumptions not supported by the Template-Based Narrative?

    \textit{Scoring guide:}
    1 indicates significant and potentially harmful fabrications.
    3 indicates minor unsupported but clinically neutral additions.
    5 indicates strict adherence to the provided source data.

    \textbf{Confidence Scoring Guide:}

    For each dimension, provide a confidence score from 0.0 to 1.0 indicating certainty of the evaluation.
    1.0 indicates complete certainty.
    0.8--0.9 indicates high confidence.
    0.6--0.7 indicates moderate confidence.
    0.4--0.5 indicates low confidence.
    0.1--0.3 indicates very low confidence.
    0.0 indicates no confidence.

    Return your evaluation result in the following JSON format:

    \texttt{\{"scores": [...], "confidence": [...], "critique": \{...\}\}}
    \\
    \hline
    \caption{\textbf{Prompt template for LLM-based narrative quality evaluation.}
      This prompt implements an LLM-as-a-Judge framework to assess the quality and safety of rewritten narratives.
      Judge models compare template-based and enhanced narratives and assign scores across five dimensions,
    along with confidence estimates and a brief rationale.}
    \label{tab:template_evaluation_quality}
  \end{longtable}
  \twocolumn

  \begin{table}[H]
    \centering
    \small
    \begin{tabularx}{\columnwidth}{|X|}
      \hline
      \rowcolor{gray!15}\textbf{System Prompt} \\
      \hline
      You are a clinical evaluation model. Your task is to extract symptom information from two texts:
      a ground-truth reference summary and a model-generated prediction summary.
      Do not interpret or rewrite either text.
      Do not generate explanations or narrative.
      You must only evaluate whether symptoms are present and how severe they are.

      You must evaluate the following 14 symptom categories:
      \begin{enumerate}[noitemsep,topsep=2pt,leftmargin=1.5em]
        \item Anhedonia (loss of interest or pleasure)
        \item DepressedMood
        \item SleepDisturbance
        \item FatigueEnergy
        \item AppetiteChange
        \item SelfWorthGuilt
        \item Concentration
        \item PsychomotorChange
        \item SuicidalIdeation
        \item SomaticDiscomfort
        \item AnxietyArousal
        \item UncontrollableWorry
        \item NegativeEvent
        \item OverallSeverity
      \end{enumerate}

      Severity scale is ordinal and must be inferred from the overall semantic strength of the description.
      If a symptom is not present in a text, you must set both presence and severity to 0.
      \\
      \hline
      \rowcolor{gray!15}\textbf{User Prompt Template} \\
      \hline
      \textbf{Reference Summary:}\\
      \{reference\}

      \textbf{Prediction Summary:}\\
      \{prediction\}
      \\
      \hline
      \rowcolor{gray!15}\textbf{Structured Response Schema} \\
      \hline
      \texttt{SymptomEvaluation} object with 14 symptom fields.
      Each field contains:
      \begin{itemize}[noitemsep,topsep=2pt,leftmargin=1.5em]
        \item \texttt{ref\_presence}: $\{0, 1\}$ — whether symptom is present in reference
        \item \texttt{pred\_presence}: $\{0, 1\}$ — whether symptom is present in prediction
        \item \texttt{ref\_severity}: $\{0, 1, 2, 3\}$ — severity level in reference
        \item \texttt{pred\_severity}: $\{0, 1, 2, 3\}$ — severity level in prediction
      \end{itemize}

      \vspace{0.5em}
      \textbf{Severity Scale:}
      0 = Not mentioned/None,
      1 = Mild,
      2 = Moderate,
      3 = Severe
      \\
      \hline
    \end{tabularx}
    \caption{\textbf{Prompt template for structured symptom evaluation on narratives.}
      This prompt uses a structured output schema to extract symptom presence and ordinal severity
    from both reference and prediction narratives.}
    \label{tab:structured-eval-prompt}
  \end{table}

  \begin{table}[H]
    \centering
    \small
    \begin{tabularx}{\columnwidth}{|X|}
      \hline
      \rowcolor{gray!15}\textbf{System Prompt} \\
      \hline
      You are a clinical evaluation model. Your task is to assess the severity of a symptom or behavior
      described in two texts: a ground-truth reference and a model-generated prediction.

      For \textbf{each} text, output a severity score from 0 to 3:
      \begin{itemize}[noitemsep,topsep=2pt,leftmargin=1.5em]
        \item 0: No symptom / absent / not at all
        \item 1: Mild / occasionally / somewhat
        \item 2: Moderate / often / frequently
        \item 3: Severe / almost always / very frequently
      \end{itemize}

      Base your judgment on the semantic intensity and frequency descriptors in each text.
      Do not add explanations or any additional fields.
      \\
      \hline
      \rowcolor{gray!15}\textbf{User Prompt Template} \\
      \hline
      \textbf{Question:} \{question\}

      \textbf{Reference:} \{reference\}

      \textbf{Prediction:} \{prediction\}
      \\
      \hline
      \rowcolor{gray!15}\textbf{Structured Response Schema} \\
      \hline
      \texttt{SeverityPair} object with two fields:
      \begin{itemize}[noitemsep,topsep=2pt,leftmargin=1.5em]
        \item \texttt{ref\_severity}: $\{0, 1, 2, 3\}$ — severity score for reference text
        \item \texttt{pred\_severity}: $\{0, 1, 2, 3\}$ — severity score for prediction text
      \end{itemize}
      \\
      \hline
    \end{tabularx}
    \caption{\textbf{Prompt template for QA severity evaluation.}
      This prompt uses a structured output schema to extract ordinal severity scores
    from both reference and prediction answers for single-item QA pairs.}
    \label{tab:qa-eval-prompt}
  \end{table}

  \begin{table}[H]
    \centering
    \small
    \begin{tabularx}{\columnwidth}{|X|}
      \hline
      \rowcolor{gray!15}\textbf{VLM Narrative Prompt (Image Input)} \\
      \hline
      You are a clinical psychologist interpreting behavioral and physiological data visualized in the image below. Each chart represents one data stream collected continuously during the past four hours:

      \begin{enumerate}[noitemsep,topsep=0pt]
        \item Heart rate (bpm): indicator of arousal, stress, and autonomic balance.
        \item Pseudoactigraphy: derived from wrist accelerometer signals, representing movement intensity and rest–activity rhythm.
        \item Steps per minute: reflects overall mobility and engagement in physical activity.
        \item Stress level: Garmin HRV-based estimation of physiological stress.
        \item GPS coordinates (longitude and latitude): capture spatial mobility and time spent in different environments.
        \item Phone unlock status: number of unlock events per minute, representing cognitive or social engagement.
      \end{enumerate}

      Additional contextual features:\\
      \{contextual\_features\}

      \textbf{Your task:}\\
      Analyze the figure carefully and provide a clinical summary of the user's recent psychological and behavioral state, as if you were writing a short report based on PHQ-9–related observations.

      In your reasoning, consider the following aspects:
      \begin{itemize}[noitemsep,topsep=0pt]
        \item Interest or pleasure in activities
        \item Mood (sadness, hopelessness)
        \item Energy and fatigue
        \item Sleep quality
        \item Appetite or eating changes
        \item Self-evaluation (guilt, self-worth)
        \item Concentration or attention
        \item Psychomotor changes (slowed or restless behavior)
        \item Anxiety or worry
      \end{itemize}

      \textbf{Important Instructions:}
      \begin{itemize}[noitemsep,topsep=0pt]
        \item Write a single, concise clinical narrative (150--200 words maximum)
        \item Do NOT generate multiple versions, drafts, or revisions
        \item Avoid mentioning charts, axes, or numerical values
        \item Write naturally, as if summarizing for another clinician
        \item Stop immediately after completing the summary
      \end{itemize}
      \\
      \hline
    \end{tabularx}
    \caption{\textbf{Prompt template for VLM-based narrative generation.} The model receives a multi-panel time-series visualization and contextual features (sleep duration, conversation length) to generate a clinical mental health summary.}
    \label{tab:vlm-narrative-prompt}
  \end{table}

  \begin{table}[H]
    \centering
    \small
    \begin{tabularx}{\columnwidth}{|X|}
      \hline
      \rowcolor{gray!15}\textbf{VLM QA Prompt (Image Input)} \\
      \hline
      You are a clinical psychologist interpreting behavioral and physiological data shown in the accompanying image. The visualization contains seven streams collected across the past four hours:

      \begin{enumerate}[noitemsep,topsep=0pt]
        \item Heart rate (bpm): indicator of arousal, stress, and autonomic balance.
        \item Pseudoactigraphy: derived from wrist accelerometer signals, representing movement intensity and rest–activity rhythm.
        \item Steps per minute: reflects overall mobility and engagement in physical activity.
        \item Stress level: Garmin HRV-based estimation of physiological stress.
        \item GPS coordinates (longitude and latitude): capture spatial mobility and time spent in different environments.
        \item Phone unlock status: number of unlock events per minute, representing cognitive or social engagement.
      \end{enumerate}

      Additional contextual features:\\
      \{contextual\_features\}

      \textbf{Question:}\\
      \{question\}

      \textbf{Your task:}\\
      Analyze the figure and answer the question directly. Base your reasoning only on observable behavioral and physiological patterns plus the contextual features. Produce a concise, clinically grounded answer (2--3 sentences) with no bullet points.

      \textbf{Important Instructions:}
      \begin{itemize}[noitemsep,topsep=0pt]
        \item Provide one clear answer, no alternative scenarios
        \item Avoid referencing charts, axes, or specific numeric values
        \item Do not speculate beyond the available evidence
        \item Stop immediately after giving the answer
      \end{itemize}
      \\
      \hline
    \end{tabularx}
    \caption{\textbf{Prompt template for VLM-based question answering.} The model receives a time-series visualization, contextual features, and a clinical question to generate a focused answer.}
    \label{tab:vlm-qa-prompt}
  \end{table}

  \begin{table}[H]
    \centering
    \small
    \begin{tabularx}{\columnwidth}{|X|}
      \hline
      \rowcolor{gray!15}\textbf{Text Baseline Narrative Prompt} \\
      \hline
      You are a clinical reasoning assistant that interprets physiological and behavioral time-series data to infer a user's psychological and physical wellbeing.

      You will receive seven time-series streams recorded over the last 4 hours, each represented in text form, along with two summary variables (sleep duration and conversation length).

      \textbf{Time-series Inputs:}
      \begin{enumerate}[noitemsep,topsep=0pt]
        \item Heart rate (1 reading per minute, length 1440) \texttt{<ts></ts>}
        \item Pseudoactigraphy (accelerometer-based movement intensity $\times$ zero-crossing rate, length 480) \texttt{<ts></ts>}
        \item Steps per minute (length 240) \texttt{<ts></ts>}
        \item Stress level (length 240) \texttt{<ts></ts>}
        \item GPS longitude (length 24) \texttt{<ts></ts>}
        \item GPS latitude (length 24) \texttt{<ts></ts>}
        \item Phone unlock status (binary 0/1 per minute, length 240) \texttt{<ts></ts>}
      \end{enumerate}

      \{sleep\_conversation\}

      \textbf{Task:}\\
      Using only the provided textual data, produce a short clinical summary (about one concise paragraph) describing the user's psychological and physical state over the last 4 hours.

      Your description should resemble a human-written mental-health assessment and cover these symptom dimensions:
      \begin{itemize}[noitemsep,topsep=0pt]
        \item Interest or pleasure in activities
        \item Depressed or hopeless mood
        \item Sleep quality or restfulness
        \item Energy or fatigue
        \item Appetite or eating pattern
        \item Self-esteem or self-criticism
        \item Concentration or focus
        \item Psychomotor activity (slowed or restless)
        \item Thoughts of self-harm or hopelessness
        \item Physical discomfort (e.g., headache, stomach, or body aches)
        \item Nervousness or anxiety
        \item Uncontrollable worry
        \item Exposure to recent negative events
      \end{itemize}

      \textbf{Output Format:}\\
      Return only the narrative summary paragraph. Do not include bullet points, lists, or section headers. End with a brief statement summarizing the likely mood severity (e.g., mild, moderate, or severe depression/anxiety).
      \\
      \hline
    \end{tabularx}
    \caption{\textbf{Prompt template for text-based narrative generation.} The \texttt{<ts></ts>} placeholders are replaced with raw numerical time-series arrays at inference time.}
    \label{tab:text-narrative-prompt}
  \end{table}

  \begin{table}[H]
    \centering
    \small
    \begin{tabularx}{\columnwidth}{|X|}
      \hline
      \rowcolor{gray!15}\textbf{Text Baseline QA Prompt} \\
      \hline
      You are a clinical reasoning assistant that interprets physiological and behavioral time-series data to answer clinical wellbeing questions about the user.

      You will receive seven time-series streams recorded over the last 4 hours, each represented in text form, along with two summary variables (sleep duration and conversation length).

      \textbf{Time-series Inputs:}
      \begin{enumerate}[noitemsep,topsep=0pt]
        \item Heart rate (1 reading per minute, length 1440) \texttt{<ts></ts>}
        \item Pseudoactigraphy (accelerometer-based movement intensity $\times$ zero-crossing rate, length 480) \texttt{<ts></ts>}
        \item Steps per minute (length 240) \texttt{<ts></ts>}
        \item Stress level (length 240) \texttt{<ts></ts>}
        \item GPS longitude (length 24) \texttt{<ts></ts>}
        \item GPS latitude (length 24) \texttt{<ts></ts>}
        \item Phone unlock status (binary 0/1 per minute, length 240) \texttt{<ts></ts>}
      \end{enumerate}

      \{sleep\_conversation\}

      \textbf{Question:}\\
      \{question\}

      \textbf{Answer Requirements:}
      \begin{itemize}[noitemsep,topsep=0pt]
        \item Provide a concise, clinically grounded answer in one or two sentences.
        \item Refer only to the information implied by the time-series data; do not add external facts.
        \item If the data is insufficient, explicitly say so.
      \end{itemize}

      \textbf{Answer:}
      \\
      \hline
    \end{tabularx}
    \caption{\textbf{Prompt template for text-based question answering.} The \texttt{<ts></ts>} placeholders are replaced with raw numerical time-series arrays at inference time.}
    \label{tab:text-qa-prompt}
  \end{table}

  \begin{table}[H]
    \centering
    \small
    \begin{tabularx}{\columnwidth}{X}
      \toprule
      \textbf{Zero-shot Baseline Output Example} \\
      \midrule
      The user demonstrates a pattern of moderate physical activity, with periods of increased movement and heart rate, suggesting engagement in various tasks or activities. Sleep duration of 10.02 hours indicates good rest, potentially supporting mood stability. The stress level shows fluctuations, with peaks that may correlate with increased activity or cognitive demands, as evidenced by frequent phone unlocking. The GPS data reveals limited spatial mobility, possibly reflecting a routine or sedentary lifestyle. The conversation length of 2303 seconds suggests active social engagement, which can be beneficial for mood. Overall, the user appears to have a balanced routine with adequate rest and social interaction, though the fluctuating stress levels warrant further investigation to assess potential underlying anxiety or worry. The data does not strongly indicate significant mood disturbances or psychomotor changes, but continued monitoring is advised to capture any emerging patterns related to depression or anxiety. \\
      \bottomrule
    \end{tabularx}
    \caption{\textbf{Example output from a zero-shot baseline model.} Without few-shot guidance, the model focuses on descriptive interpretations of sensor data and behavioral signals, rather than producing a PHQ-9--style narrative summary aligned with EMA-derived symptom categories and severity framing.}
    \label{tab:zero_shot_example}
  \end{table}

  \end{document}